%% file: main.tex
\documentclass[runningheads]{llncs}

\PassOptionsToPackage{table}{xcolor}

\usepackage{eccv}

\usepackage{eccvabbrv}
\usepackage{graphicx}
\usepackage{array}
\usepackage{booktabs}
\usepackage{multirow}
\usepackage{wrapfig}
\usepackage{pifont}
\usepackage{tikz}
\usepackage{placeins}

\input{math_commands.tex}

\usepackage[breaklinks,colorlinks,citecolor=eccvblue,urlcolor=eccvblue]{hyperref}

\providecommand{\citep}[1]{\cite{#1}}

\begin{document}

\title{Lumina-OmniLV: A Unified Multimodal Framework for General Low-Level Vision}
\titlerunning{Lumina-OmniLV}

\author{Yuandong Pu\inst{1,2} \and Le Zhuo\inst{2} \and Kaiwen Zhu\inst{1,2} \and
Liangbin Xie\inst{3,4} \and Wenlong Zhang\inst{2} \and Xiangyu Chen\inst{2,6} \and
Peng Gao\inst{2} \and Yu Qiao\inst{2} \and Chao Dong\inst{4,5,2} \and
Yihao Liu\inst{2}\thanks{Corresponding author.}}
\authorrunning{Y. Pu et al.}
\institute{Shanghai Jiao Tong University \and
Shanghai AI Laboratory \and
University of Macau \and
Shenzhen Institutes of Advanced Technology, Chinese Academy of Sciences \and
Shenzhen University of Advanced Technology \and
Institute of Artificial Intelligence (TeleAI), China Telecom}

\maketitle

\input{sec/0_abstract}
\input{sec/1_intro}
\input{sec/2_related_work}
\input{sec/3_method}
\input{sec/4_experiment}
\FloatBarrier
\input{sec/5_conclusion}
\FloatBarrier
\newpage
\bibliographystyle{splncs04}
\bibliography{main}

\end{document}

%% file: math_commands.tex
\usepackage{amsmath,amsfonts,bm}

\def\eqref#1{equation~\ref{#1}}

\def\1{\bm{1}}

\DeclareMathAlphabet{\mathsfit}{\encodingdefault}{\sfdefault}{m}{sl}
\SetMathAlphabet{\mathsfit}{bold}{\encodingdefault}{\sfdefault}{bx}{n}



%% file: sec/0_abstract.tex
\begin{abstract}
We present \textbf{Lumina-OmniLV} (abbreviated as \textbf{OmniLV}), a universal multimodal multi-task framework for low-level vision that addresses over 100 sub-tasks across four major categories, including image restoration, image enhancement, weak-semantic dense prediction, and stylization. OmniLV leverages both textual and visual prompts to offer flexible, user-friendly interactions. Built on Diffusion Transformer (DiT)-based generative priors, our framework supports arbitrary resolutions — achieving optimal performance at 1K resolution — while preserving fine-grained details and high fidelity. Through extensive experiments, we demonstrate that separately encoding text and visual instructions, combined with co-training using shallow feature control, is essential to mitigate task ambiguity and enhance multi-task generalization. Our findings also reveal that integrating high-level generative tasks into low-level vision models can compromise detail-sensitive restoration. These insights pave the way for more robust and generalizable low-level vision systems.
\par\smallskip\noindent\textbf{Project page:} \url{https://andrew0613.github.io/OmniLV_page/}
\end{abstract}

%% file: sec/1_intro.tex
\vspace{-1em}
\section{Introduction}
\label{sec:intro}
\begin{figure}[h]
    \centering
    \includegraphics[width=.95\linewidth]{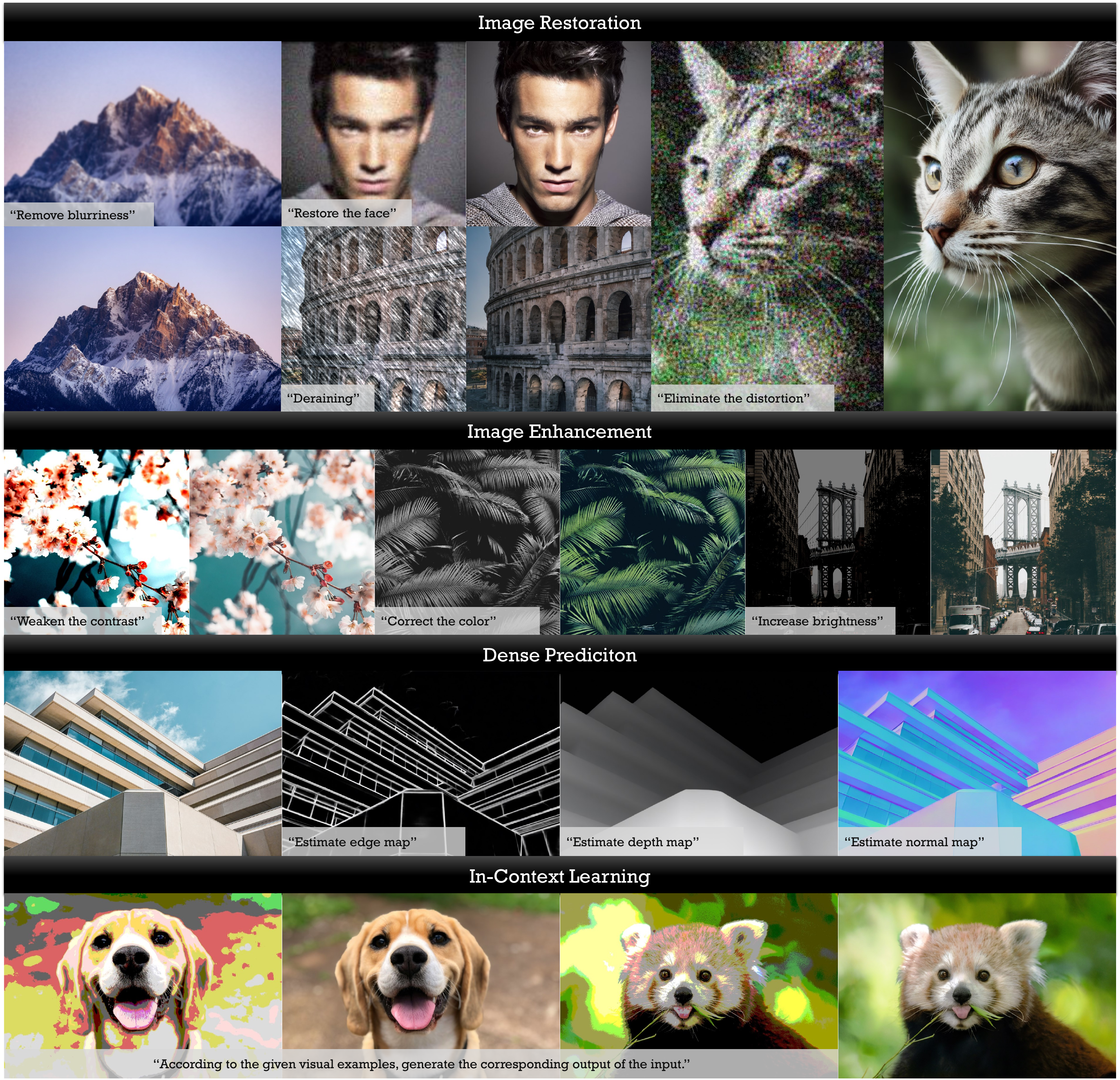}
    \caption{\textbf{Illustration of OmniLV's versatile capabilities.} As a universal framework, OmniLV handles a wide variety of low-level vision tasks within a single model that adapts to diverse input--output domains and generates high-fidelity results.}
    \label{fig:teaser}
    \vspace{-2em}
\end{figure}

The rapid evolution of large-scale foundation models has revolutionized artificial intelligence, demonstrating remarkable generalization and multi-task capabilities across various domains. Unified frameworks such as GPT-4V~\citep{achiam2023gpt}, InternVL~\citep{chen2024internvl,chen2024expanding,chen2024far}, Flamingo~\citep{alayrac2022flamingo}, OmniGen~\citep{xiao2024omnigen, wu2025omnigen2}, and Bagel~\citep{deng2025emerging} have showcased impressive performance by leveraging large-scale pretraining on multimodal datasets. These models excel in semantics-driven high-level vision tasks, such as image classification, understanding, generation, and editing. In contrast, unified models for low-level vision remain fragmented and underexplored.

Low-level vision encompasses a broad spectrum of tasks, including image restoration~\citep{dong2015image,zhang2017beyond, chen2024comparative, chen2023activating,lin2024diffbir, yu2024scaling, hypir}, image enhancement~\citep{cai2023retinexformer,Zamir2020MIRNet, chen2021new, chen2021hdrunet,wang2019underexposed}, style transfer~\citep{gatys2016image,huang2017arbitrary}, and weak-semantic dense prediction~\citep{yang2024depth, kirillov2023segment, ravi2024sam} (e.g., edge detection, depth estimation, normal map estimation). Unlike high-level vision tasks that rely on predefined semantic understanding, most low-level vision tasks do not require explicit object-level reasoning. Instead, they focus on pixel-level fidelity, fine-grained texture reconstruction, and feature extraction. This distinction makes the unification of low-level vision tasks particularly challenging, as different tasks often operate in vastly different output domains.

Existing approaches to low-level vision remain limited in generalization, usability, and scalability. Task-specific models~\citep{chen2023activating, li2022all} are designed to handle a single task (e.g., denoising, deblurring, super-resolution), requiring extensive redesign and retraining to adapt to new tasks. All-in-one restoration models, such as AirNet~\citep{li2022all}, PromptIR~\citep{potlapalli2023promptir}, and OneRestore~\citep{guo2024onerestore}, integrate multiple restoration tasks within a single framework, yet remain restricted to in-domain restoration, unable to generalize to cross-domain tasks such as feature extraction or style transfer. Visual-prompt-based models, such as PromptGIP~\citep{liu2023unifying} and GenLV~\citep{chen2024learning, chen2025exploring}, extend to cross-domain tasks using image prompt pairs, but require carefully crafted prompts, making them less intuitive compared to text-driven interaction. Furthermore, many existing methods operate only on fixed-resolution images, severely limiting their flexibility and real-world applicability. To summarize, high-resolution image processing remains challenging, leaving ample room for improvement in task adaptability.

Given the complexity and diversity of low-level vision, a truly universal model must handle multiple task domains while reliably preserving fine-grained details and high fidelity. A key requirement for such a model is flexible interaction mechanisms. While text-based instructions offer a convenient and intuitive way to specify tasks (e.g., “remove noise from this image”, “enhance brightness”, and “estimate the Canny edge”), certain tasks---such as style transfer---are difficult to define using text alone. Visual prompts in the form of exemplar image pairs offer an effective alternative, allowing the model to infer complex, task-specific transformations through visual analogy. Thus, an ideal general low-level vision model should integrate both textual and visual prompts for versatile task execution.

To address these challenges, we propose \textbf{OmniLV}, a universal multimodal multi-task framework for low-level vision, capable of handling 100+ sub-tasks via both textual and visual prompts. Built on Diffusion Transformer (DiT)-based generative priors~\citep{ho2020denoising, rombach2022high,peebles2023scalable, esser2024scaling}, our model improves generalization and output quality across tasks. Unlike prior models constrained to fixed resolutions, our framework supports arbitrary resolutions and achieves strong performance at 1K resolution. We systematically explore multimodal fusion strategies and propose a simple yet effective design that prevents task misinterpretation issues. Fig.~\ref{fig:teaser} presents the versatile capabilities of OmniLV.

Throughout the development of OmniLV, we have gained several key insights that shape the design of a robust and generalizable low-level vision model. First, we find that separately encoding text-based and visual instructions is crucial for preventing task ambiguity, as naive fusion can lead to task misinterpretations (Sec. \ref{sec:multimodel}).  Additionally, co-training the base model with shallow feature control proves to be an effective strategy for enhancing multi-task generalization (Sec. \ref{sec:condition}). Furthermore, incorporating high-level generative or editing tasks into a low-level vision model significantly compromises fidelity, particularly in detail-sensitive restoration tasks (Sec. \ref{sec:exploration}). These findings highlight the need for dedicated multimodal architectures tailored for low-level vision tasks.

In summary, our contributions are threefold. (1) We propose OmniLV, a unified multimodal framework that covers four major low-level vision categories (100+ sub-tasks) via both text instructions and in-context visual exemplars. (2) We systematically analyze multimodal encoding and conditioning designs and identify simple, effective choices---separate encoders and shallow feature control co-training---that improve task alignment and multi-task generalization (Sec.~\ref{sec:multimodel}). (3) We provide empirical insights into key trade-offs when building low-level vision generalists, including when incorporating high-semantic generative/editing tasks can degrade fidelity-critical restoration (Sec.~\ref{sec:exploration}).

%% file: sec/2_related_work.tex
\vspace{-1.5em}
\section{Related Work}
\vspace{-0.5em}
\subsection{Image Restoration with Generative Prior}
\vspace{-0.2em}
Diffusion-based methods have emerged as a robust framework for image restoration, converting degraded inputs into high-quality outputs through reverse denoising. Several key works illustrate the versatility of this approach~\citep{wang2024exploiting, lin2024diffbir, yang2024pixel, yu2024scaling, ai2024dreamclear, wu2024one, chen2024adversarial,yue2024arbitrary}.
StableSR~\citep{wang2024exploiting} leverages the generative priors of pre-trained text-to-image diffusion models for blind super-resolution, employing a time-aware encoder and feature wrapping to balance quality and fidelity while accommodating arbitrary resolutions. DiffBIR~\citep{lin2024diffbir} uses a two-stage pipeline where the first stage reduces degradations and the second stage employs a latent diffusion model to generate missing details, proving effective in denoising and face restoration.
PASD~\citep{yang2024pixel} extends the Stable Diffusion framework for realistic super-resolution and personalized stylization by integrating a pixel-aware mechanism that improves both resolution precision and style adaptability.
SUPIR~\citep{yu2024scaling} scales up large diffusion models such as StableDiffusion-XL, incorporating a trained adapter and a massive high-resolution dataset to enable text-guided, photo-realistic restoration in complex scenes.
However, the limitation of these approaches is that they are confined to image restoration tasks and cannot address other challenges in low-level vision.

\vspace{-0.5em}
\subsection{All-in-one Generative Models}
\vspace{-0.2em}
Developing all-in-one models is an exciting yet challenging pursuit. In the realm of image generation, various studies have sought to build versatile systems~\citep{wang2024emu3,le2024diffusiongenerate,xiao2024omnigen,lin2024pixwizard,han2024ace,mao2025ace++}. For example, OmniGen~\citep{xiao2024omnigen} encodes text and images into a unified tensor, utilizing causal attention for text tokens and bidirectional attention for image tokens. Pixwizard ~\citep{lin2024pixwizard} introduces task-specific embeddings for image editing and understanding, while ACE~\citep{han2024ace,mao2025ace++} offers a conditioning module that accepts diverse input images and processes them concurrently with a transformer. UniReal~\citep{chen2024unireal} employs a video generation framework that treats images as individual frames, providing a universal solution for various image generation and editing tasks.

Despite these advances, most of these approaches focus on image generation and editing, leaving universal models for low-level vision relatively unexplored. Visual prompt-based approaches~\citep{liu2023unifying, chen2024learning, chen2025exploring} tackle cross-domain tasks by utilizing pairs of image prompts. However, their dependence on meticulously crafted prompts renders them less intuitive and user-friendly compared to text-driven alternatives. Moreover, many current methods are restricted to fixed-resolution outputs, limiting their practical applicability.

%% file: sec/3_method.tex
\section{Method}
\subsection{Building OmniLV Step-by-Step}
In this section, we detail the key design choices and learned insights in developing a universal low-level vision model, outlining our step-by-step thinking process.
\subsubsection{Preliminary: Base Model Selection}
\label{sec:base-model}
Unlike most foundational image restoration models~\citep{ho2020denoising,rombach2022high,peebles2023scalable,esser2024scaling} that are trained from scratch using deterministic regression objectives, we leverage a pre-trained text-to-image diffusion model as a strong initialization. Pre-trained diffusion models~\citep{esser2024scaling, gao2024lumina-next, gao2024lumin-t2x, xie2024sana, flux2024}, trained on billions of images, offer rich visual priors that enhance generalization, support diverse resolutions and aspect ratios, and effectively capture the uncertainty inherent in multi-task image restoration. These properties contribute to a more robust and versatile low-level vision model.

For our base model, we initialize with Lumina-Next~\citep{gao2024lumina-next, gao2024lumin-t2x}, a flow-based diffusion transformer that introduces several architectural improvements over traditional DiT-based models~\citep{peebles2023scalable}, including 2D Rotary Positional Encoding, QK Normalization, and Sandwich Normalization. Additionally, Lumina-Next adopts a flow-matching formulation, improving training stability and accelerating convergence. To adapt this model for general low-level vision, we introduce a condition adapter that integrates inputs to enable effective task conditioning, which is illustrated in subsequent sections. The modified model is trained using a flow-matching loss to learn a conditional time-dependent velocity field, facilitating the transformation between noisy and clean image distributions. Please refer to the supplementary material for details of training loss.
We use Gemma-2B to encode the instruction and description prompts, matching the base model's text interface. In our setting, language primarily specifies the desired operation, while the input image provides most task evidence; thus a lightweight encoder is sufficient and keeps the conditioning pathway simple.


\subsubsection{Encoding Multimodal Information}
\label{sec:multimodel}
\begin{figure}[t]
\begin{minipage}{.49\linewidth}
    \centering
    \includegraphics[width=\linewidth]{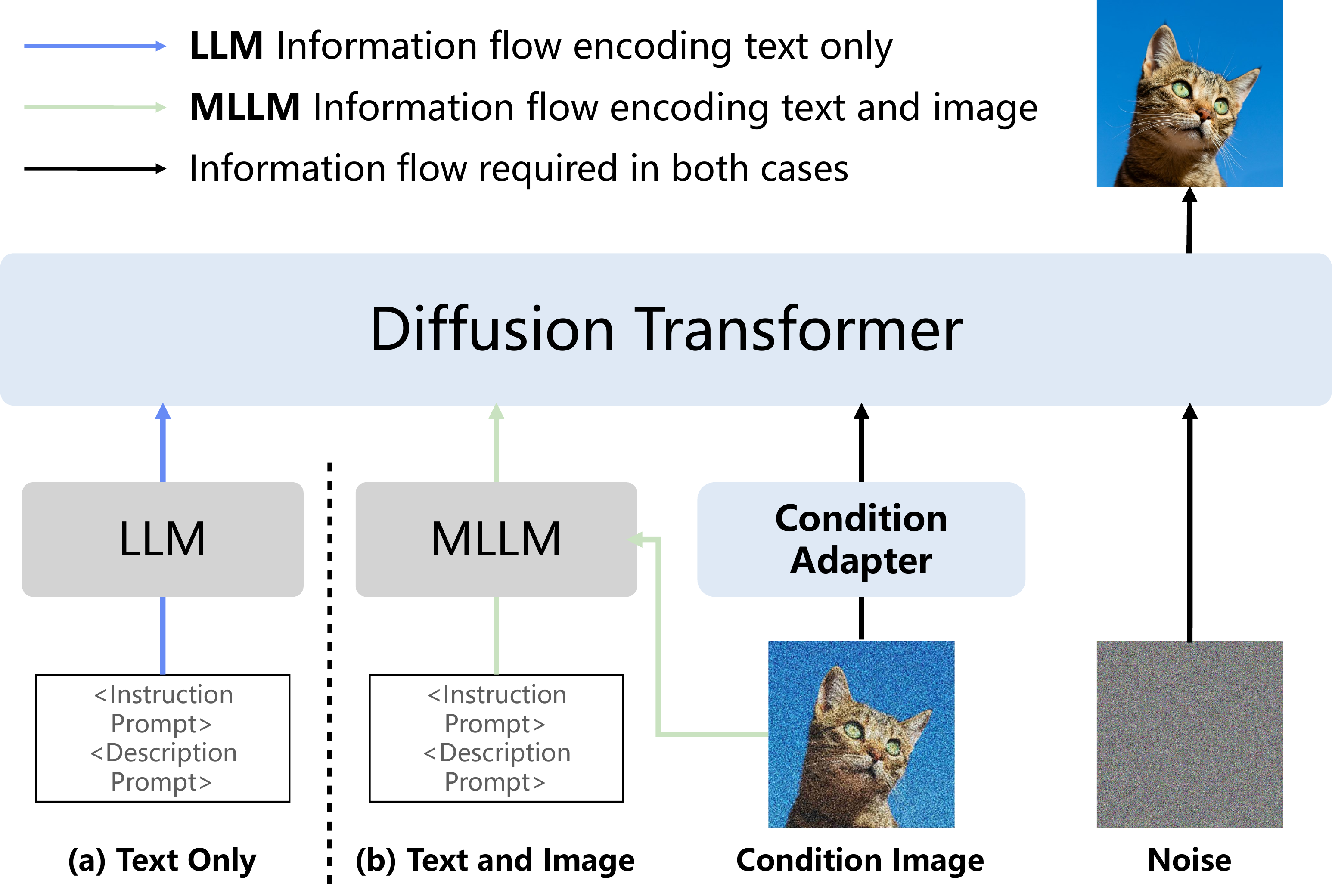}
    \captionof{figure}{Comparison between MLLM guided and LLM guided framework.}
    \label{fig:mllm_framework}
\end{minipage}
\hfill
\begin{minipage}{.49\linewidth}
\begin{minipage}{\linewidth}
\centering
\resizebox{1.0\columnwidth}{!}{
\begin{tabular}{@{}ccccccccc@{}}
\toprule
\multirow{2}{*}{}  & \multicolumn{2}{c}{Compression} & \multicolumn{2}{c}{Quantization} & \multicolumn{2}{c}{Noise} & \multicolumn{2}{c}{Inpainting} \\
\cmidrule{2-9}
                   & PSNR$\uparrow$        & MUSIQ$\uparrow$      & PSNR$\uparrow$        & MUSIQ$\uparrow$       & PSNR$\uparrow$             & MUSIQ$\uparrow$           & PSNR$\uparrow$             & MUSIQ$\uparrow$           \\ \midrule
Addition & 21.92       & \textbf{56.31}      & \textbf{18.72}       & \textbf{55.71}       & 22.34            & \textbf{59.31}           & \textbf{19.94}            & \textbf{56.96}          \\
Concat  & \textbf{21.93}       & 55.99      & 18.18       & 55.11       & \textbf{22.35}            & 57.13           & 19.86            & 55.95          \\\bottomrule
\end{tabular}}
\captionof{table}{Ablation study on whether to use addition or concatenation in in-context learning.}
\label{tab:icl}
\vspace{0.5em}
\end{minipage}
\begin{minipage}{\linewidth}
    \includegraphics[width=\linewidth]{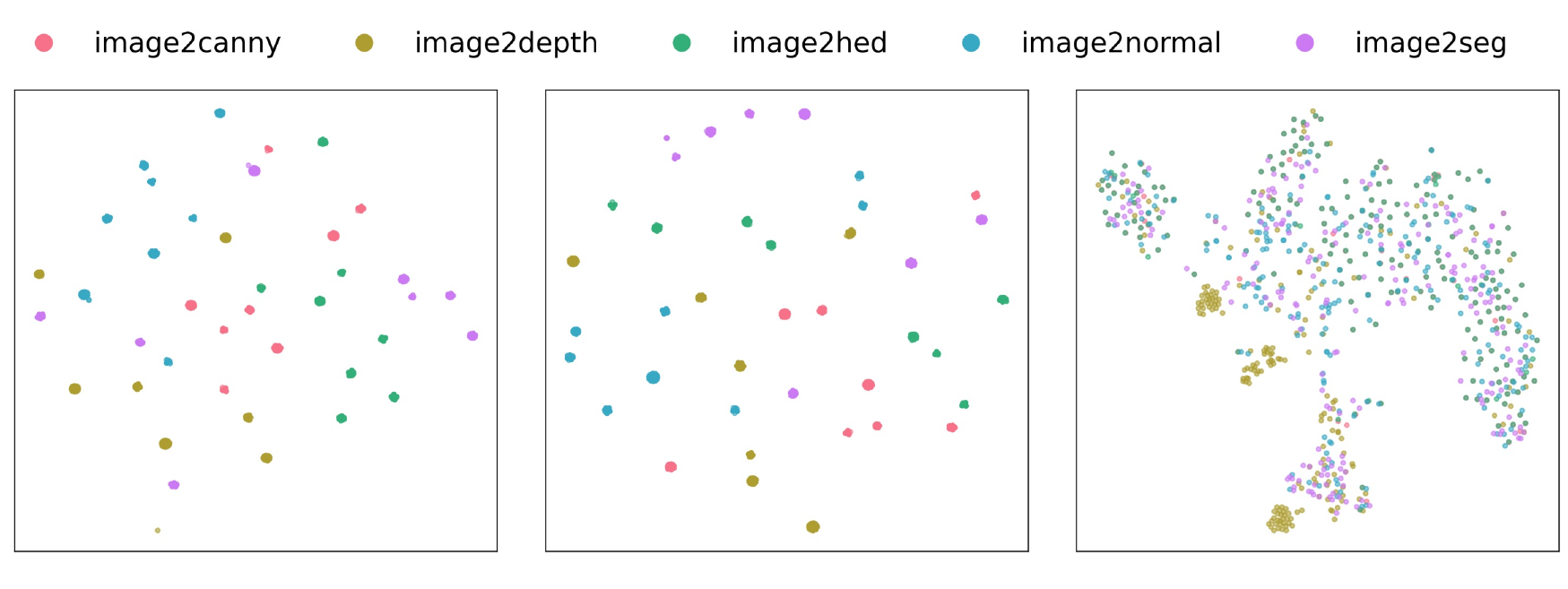}
    \captionof{figure}{t-SNE visualization of the feature space of LLM and MLLM. Each dot represents a task instruction.}
    \label{fig:task_mismatch}
\end{minipage}
\end{minipage}
\end{figure}

Given an input image $x$, our goal is to generate the target image using both textual instructions and in-context visual exemplars. We explore two different encoding strategies: (1) \textbf{Separate encoding}, where text prompts are processed using a large language model (LLM), while visual exemplars are encoded independently. (2) \textbf{Unified encoding}, where both text and visual inputs are fused within a multimodal language model (MLLM). Fig.~\ref{fig:mllm_framework} illustrates the architectural differences between these two approaches. While unified encoding benefits from parameter efficiency and leverages cross-modal correlations, we observe that it introduces critical limitations especially when applied to dense prediction tasks. Specifically, multimodal encoders often misinterpret task instructions, leading to inconsistencies in generated outputs. To better understand this issue, we visualize the encoded feature distributions in Fig.~\ref{fig:task_mismatch}. Our findings indicate that mixing text and image prompts within a single encoder leads to severe task ambiguity. Since visual tokens dominate the shared feature space, text-based instructions often get overshadowed, leading to misalignment and incorrect outputs. This is particularly problematic for dense prediction, where the desired output is tightly specified by the instruction. Details are provided in Sec.~\ref{Instruction Following Ability}.

Based on these observations, we adopt a separate encoding strategy: text instructions are processed via an LLM, while image exemplars are encoded using a visual VAE. This ensures clearer task separation, preventing interference between textual and visual guidance, and improves task accuracy across a vast number of low-level vision tasks.

\begin{figure}[t]
    \centering
    \includegraphics[width=\linewidth]{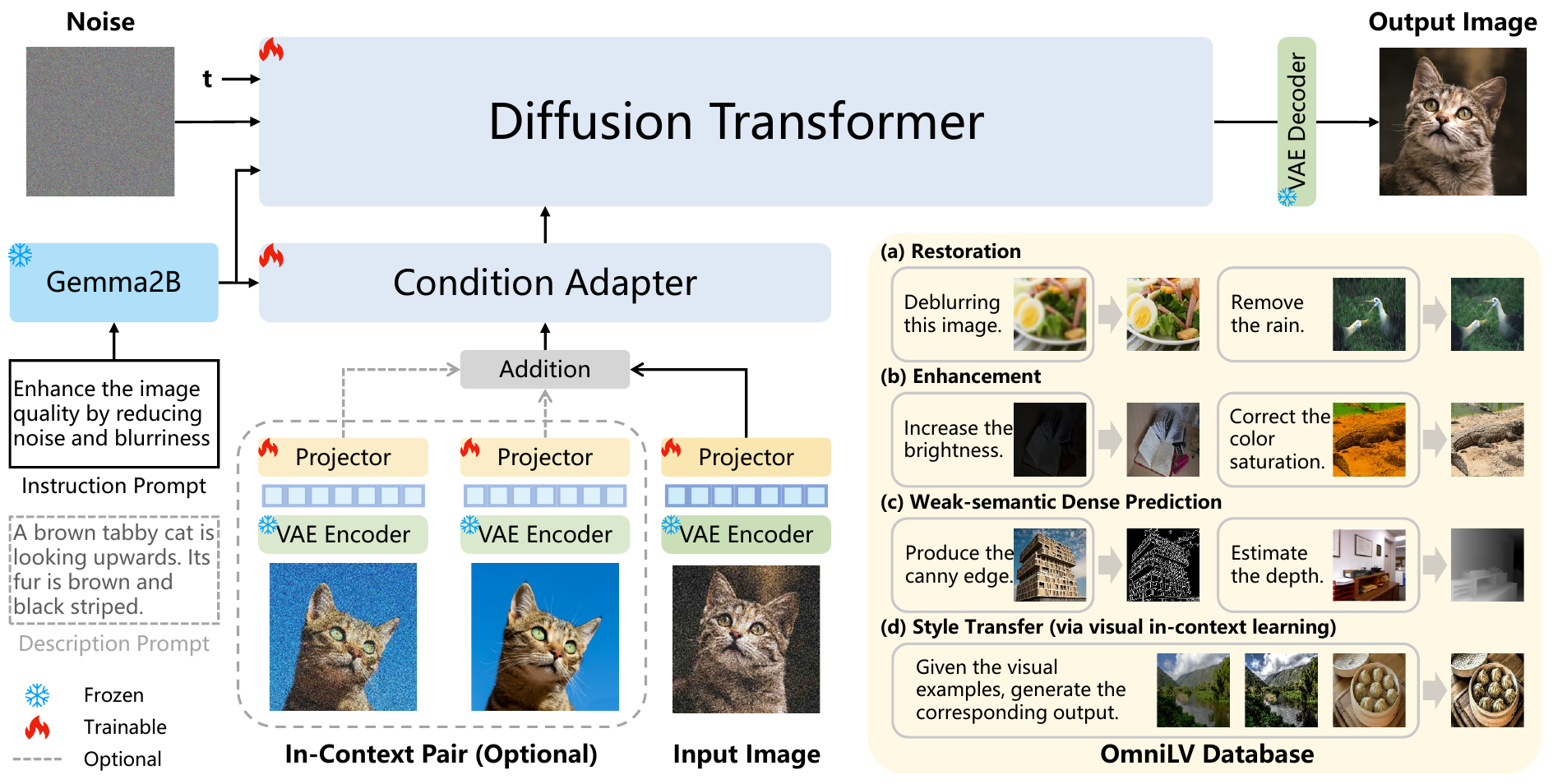}
    \caption{\textbf{Overall framework of OmniLV}. First, input images are encoded into latent space by VAE encoder. Then, we patchify the image latent and noise latent into visual tokens. Optionally, in-context pairs can be added to visual tokens to handle complex scenarios. At the same time, the instruction prompt and description prompt are processed by Gemma-2B. Finally, we decode the denoised results to get the desired output images.}
    \label{fig:framework}
\end{figure}

\subsubsection{Design Choices of Condition Integration}
\label{sec:condition}

\begin{figure}[t]
\begin{minipage}{.45\linewidth}
\centering
\includegraphics[width=\linewidth]{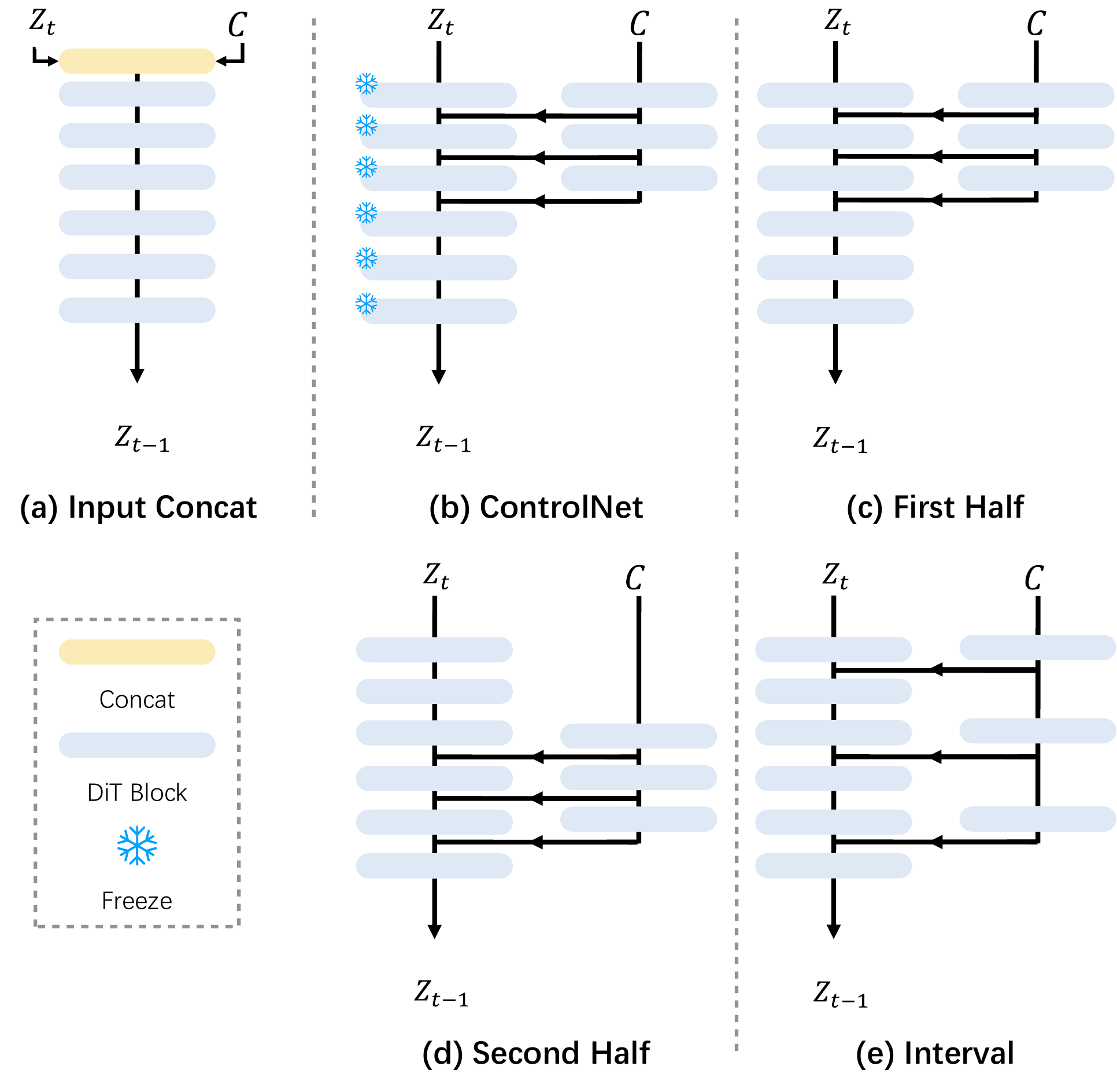}
\captionof{figure}{Schematic structures of five different variants.}
\label{fig:condition_framework}
\end{minipage}
\hfill
\begin{minipage}{.54\linewidth}
\begin{minipage}{\linewidth}
\centering
\captionof{table}{Ablation study on condition integration.}
\label{tab:condition_performance}
\scriptsize  
\setlength{\tabcolsep}{3pt}  
\resizebox{1.0\columnwidth}{!}{  
\begin{tabular}{llccccccc}
\toprule
& \multirow{2}{*}{Position} & \multirow{2}{*}{Train DM?} & \multicolumn{2}{c}{SIDD} & \multicolumn{2}{c}{RealBlurJ} & \multicolumn{2}{c}{SR} \\
\cmidrule(lr){4-5} \cmidrule(lr){6-7} \cmidrule(lr){8-9} 
                 &   &                             & PSNR$\uparrow$  & MUSIQ$\uparrow$   & PSNR$\uparrow$  & MUSIQ$\uparrow$   & PSNR$\uparrow$  & MUSIQ$\uparrow$  \\
\midrule
(a) & Input       & \ding{51}                & 32.40 & 21.91  & 22.98 & 51.66  & 22.89 & \textbf{56.89} \\
(b) & First Half         & \ding{55}                  & 25.52 & 23.53  & 22.28 & 44.41  & 21.11 & 50.44 \\
(c) & First Half         & \ding{51}                  & \textbf{34.09} & \textbf{23.96}  & \textbf{24.05} & \textbf{57.42}  & \textbf{22.93} & 56.72 \\
(d) & Second Half        & \ding{51}                & 29.60 & 23.38  & 23.15 & 54.35  & 22.96 & 56.60 \\
(e) & Interval           & \ding{51}                  & 34.07 & 23.06  & 22.77 & 53.50  & 22.80 & 56.38 \\
\bottomrule
\end{tabular}
}
\end{minipage}
\begin{minipage}{\linewidth}
\vspace{0.5em}
\centering
\captionof{table}{Effects of Various Prompt Formats. Our approach supports text prompts, visual prompts, or a combination of both.}
\label{tab:text_visual}
\resizebox{1.0\columnwidth}{!}{
\begin{tabular}{@{}cccccccccc@{}}
\toprule
\multirow{2}{*}{Text} & \multirow{2}{*}{Visual} & \multicolumn{2}{c}{Blur} & \multicolumn{2}{c}{Noise} & \multicolumn{2}{c}{Contrast Adju.} & \multicolumn{2}{c}{Saturate Adju.} \\
\cmidrule{3-10}
                            &                               & PSNR$\uparrow$        & MUSIQ$\uparrow$      & PSNR$\uparrow$        & MUSIQ$\uparrow$       & PSNR$\uparrow$             & MUSIQ$\uparrow$           & PSNR$\uparrow$             & MUSIQ$\uparrow$           \\ \midrule
\ding{51}                         & \ding{55}                            & \textbf{22.57}       & 68.95      & \textbf{23.53}       & 69.23       & \textbf{20.90}            & 69.95           & \textbf{21.79}            & 70.90           \\
\ding{55}                          & \ding{51}                           & 21.99       & 67.59      & 22.71       & 67.66       & 20.09            & 66.58           & 20.14            & 68.08           \\
\ding{51}                         & \ding{51}                           & 22.50       & \textbf{68.99}      & 23.07       & \textbf{69.37}       & 20.50            & \textbf{70.09}           & 21.00            & \textbf{70.91}           \\ \bottomrule
\end{tabular}}
\end{minipage}
\begin{minipage}{\linewidth}
\vspace{0.5em}
\centering
\captionof{table}{Ablation study for the training data.}
\label{tab:relationship}
\resizebox{1.0\columnwidth}{!}{
\begin{tabular}{@{}ccccccccc@{}}
\toprule
\multirow{2}{*}{High-semantic?}  & \multicolumn{2}{c}{Blur} & \multicolumn{2}{c}{Noise} & \multicolumn{2}{c}{Contrast Adju.} & \multicolumn{2}{c}{Saturate Adju.} \\
\cmidrule{2-9}
                   & PSNR$\uparrow$        & MUSIQ$\uparrow$      & PSNR$\uparrow$        & MUSIQ$\uparrow$       & PSNR$\uparrow$             & MUSIQ$\uparrow$           & PSNR$\uparrow$             & MUSIQ$\uparrow$           \\ \midrule
\ding{55} & \textbf{21.13}       & \textbf{58.78}      & \textbf{22.34}       & 57.39       & \textbf{19.03}            & \textbf{56.95}           & \textbf{18.64}            & \textbf{56.60}           \\
\ding{51}  & 20.97       & 57.06      & 22.18       & \textbf{59.31}       & 18.76            & 55.80           & 18.58            & 56.55          \\\bottomrule
\end{tabular}}
\end{minipage}
\end{minipage}
\end{figure}

Integrating condition images into diffusion models is commonly achieved through two primary approaches: (1) \textbf{Feature Injection:} A trainable adapter injects feature maps into a frozen diffusion model~\citep{zhang2023adding, mou2024t2i}. (2) \textbf{Input Concatenation:} Condition images are concatenated with inputs, and the entire model is fine-tuned. These designs have been widely used in in-domain single task (e.g. image restoration, canny2image), achieving remarkable results \citep{yu2024scaling, lin2024diffbir}. To systematically investigate condition integration strategies for general low-level vision tasks, we conduct comparative experiments evaluating different design choices (see Fig.~\ref{fig:condition_framework}). Our findings, summarized in Tab. \ref{tab:condition_performance}, are as follows: (1) Training only the adapter is suboptimal (settings (b) \& (c)), indicating that fine-tuning the base model is necessary for adapting generative priors to diverse low-level tasks. (2) While input concatenation is efficient(setting(a)), adding additional parameters to process the condition image enhances performance (setting (c)), suggesting that explicitly modeling condition images helps extract more relevant structural and contextual information. (3) The injection position significantly influences the performance (settings (c), (d), \& (e)). Integrating condition information in the first half of the network leads to better results, likely because early-stage modulation ensures stronger feature guidance throughout the process.

Based on these findings, we propose a co-training condition adapter, which jointly optimizes the adapter and base model. Unlike ControlNet-like architectures, which keep the base model frozen, our approach ensures deeper feature alignment, improving multi-task generalization and fidelity.

\subsubsection{Enabling In-Context Learning}
\label{sec:icl}

While text prompts can effectively guide tasks, many low-level vision tasks (e.g., stylization) require precise visual instructions that are difficult to express linguistically. To address this, we compare two paradigms for visual prompt integration: (1) \textbf{Input Concatenation}~\citep{chen2024unireal, xiao2024omnigen, wang2024lavin}, where visual prompts are concatenated along the token dimension:
\begin{equation}
\mathbf{H}_{\text{fused}} = [\mathbf{H}_{\text{img}};\ \mathbf{H}_{\text{prompt}_1};\ \dots;\ \mathbf{H}_{\text{prompt}_n}],
\end{equation} where $\mathbf{H}_{\text{img}}$ and $\mathbf{H}_{\text{prompt}_i}$ denote latent representation of input image and latent representation of i-th visual prompt, and  $\mathbf{H}_{\text{fused}}$ denotes the combined latent representation. (2) \textbf{Projection-Addition}~\cite{wang2023context}, which employs lightweight projectors to align visual prompts with the latent space before summation:
\begin{equation}
\mathbf{H}_\text{fused} = \mathbf{H}_\text{image} + \sum_{i=1}^n \phi_i(\mathbf{H}_{\text{prompt}_i}),
\end{equation}
where $\phi(\cdot)$  denotes linear projectors. Fig.~\ref{fig:framework} illustrates the architectural differences between these two approaches, where the concatenation method can be seen as a variation where the ``Projector-Addition'' module is replaced with a Concatenation operation.
Tab. \ref{tab:icl} presents the quantitative comparison, demonstrating that projection-addition outperforms input concatenation across different tasks. This suggests that projection-based alignment better preserves task-relevant information.

\noindent\textbf{Final Architecture.}
Based on these insights, we design the final architecture of OmniLV, as illustrated in Fig.~\ref{fig:framework}. Our approach unifies diverse low-level vision tasks while ensuring strong multimodal conditioning and in-context learning capabilities.

\definecolor{RestorationMain}{RGB}{220,91,83}
\definecolor{EnhancementMain}{RGB}{245,182,53}
\definecolor{AnnotationMain}{RGB}{76,137,237}
\definecolor{StylizationMain}{RGB}{73,168,100}

\definecolor{RestorationSub1}{RGB}{247,184,179}  
\definecolor{RestorationSub2}{RGB}{244,178,173}  
\definecolor{RestorationSub3}{RGB}{242,172,167}  
\definecolor{RestorationSub4}{RGB}{240,166,161}  
\definecolor{RestorationSub5}{RGB}{235,154,149}  
\definecolor{RestorationSub6}{RGB}{233,149,144}  
\definecolor{RestorationSub7}{RGB}{231,143,138}  
\definecolor{RestorationSub8}{RGB}{229,137,132}  
\definecolor{RestorationSub9}{RGB}{226,131,126}  
\definecolor{RestorationSub10}{RGB}{224,125,120}  
\definecolor{RestorationSub11}{RGB}{222,119,114}  

\definecolor{EnhancementSub1}{RGB}{255,218,63}  
\definecolor{EnhancementSub2}{RGB}{249,211,61}  
\definecolor{EnhancementSub3}{RGB}{243,205,59}  
\definecolor{EnhancementSub4}{RGB}{238,199,57}  
\definecolor{EnhancementSub5}{RGB}{232,195,56}  
\definecolor{EnhancementSub6}{RGB}{227,190,54}  
\definecolor{EnhancementSub7}{RGB}{221,181,52}  
\definecolor{EnhancementSub8}{RGB}{215,175,50}  
\definecolor{EnhancementSub9}{RGB}{210,169,48}  
\definecolor{EnhancementSub10}{RGB}{204,163,47} 
\definecolor{EnhancementSub11}{RGB}{199,157,45}  

\definecolor{AnnotationSub1}{RGB}{178,196,233}  
\definecolor{AnnotationSub2}{RGB}{165,188,234}  
\definecolor{AnnotationSub3}{RGB}{152,180,235}  
\definecolor{AnnotationSub4}{RGB}{139,172,236}  
\definecolor{AnnotationSub5}{RGB}{126,164,237}  
\definecolor{AnnotationSub6}{RGB}{113,156,238}  

\definecolor{StylizationSub1}{RGB}{87,201,120}  
\definecolor{StylizationSub2}{RGB}{79,184,110}  
\definecolor{StylizationSub3}{RGB}{72,167,100}  
\definecolor{StylizationSub4}{RGB}{65,150,90}   
\definecolor{StylizationSub5}{RGB}{58,133,80}   
\definecolor{StylizationSub6}{RGB}{51,116,70}   

\begin{figure}[!t]
	    \centering
	    \begin{minipage}{.37\linewidth}
	        \centering
	        \includegraphics[width=0.9\linewidth]{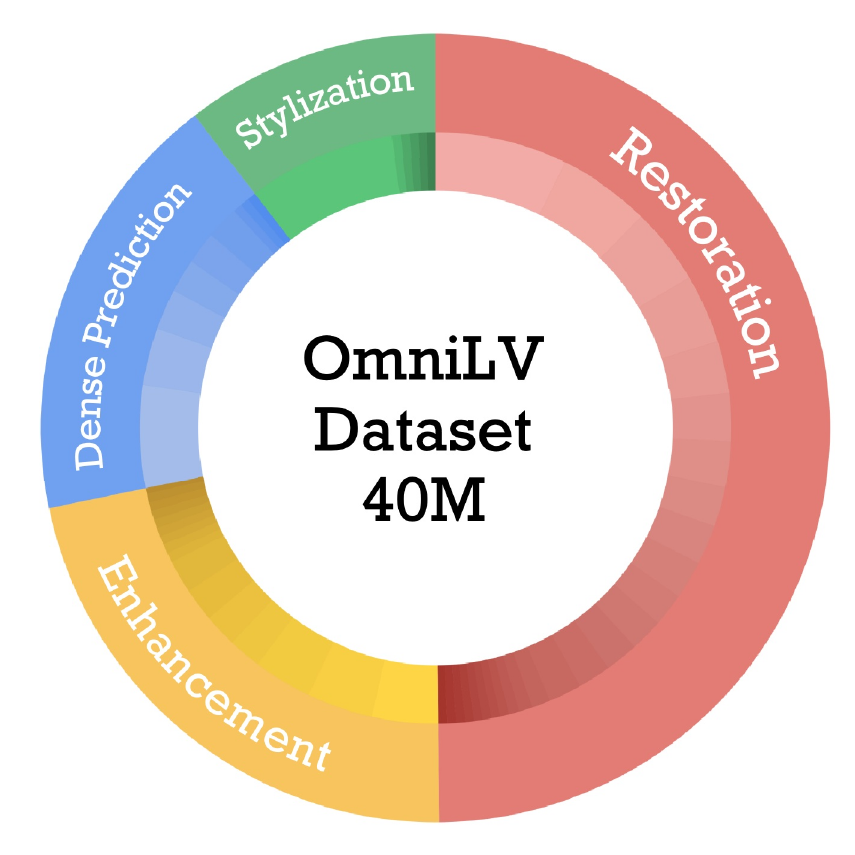}
	    \end{minipage}%
	    \hfill
	    \begin{minipage}{.61\linewidth}
	        \centering
	        \renewcommand{\arraystretch}{1.2}
	        \setlength\tabcolsep{2pt}
	        \scriptsize
        
	        \resizebox{\linewidth}{!}{%
	        \begin{tabular}{@{}>{\raggedright\arraybackslash}p{2cm}>{\raggedright\arraybackslash}p{1.8cm}>{\raggedright\arraybackslash}p{1.8cm}>{\raggedright\arraybackslash}p{1.8cm}@{}}
            \cellcolor{RestorationMain}\textcolor{white}{Restoration (49.7\%)} &
            \tikz[baseline=0.05em] \fill [RestorationSub1] (0,0) rectangle (0.6em,0.6em); Mix Deg. (7.2\%) &
            \tikz[baseline=0.05em] \fill [RestorationSub2] (0,0) rectangle (0.6em,0.6em); Face Res. (5.1\%) &
            \tikz[baseline=0.05em] \fill [RestorationSub3] (0,0) rectangle (0.6em,0.6em); SR (4.2\%) \\[0.3em]

            \tikz[baseline=0.05em] \fill [RestorationSub4] (0,0) rectangle (0.6em,0.6em); Inpainting (3.7\%) & 
            \tikz[baseline=0.05em] \fill [RestorationSub5] (0,0) rectangle (0.6em,0.6em); Blur (2.9\%) &
            \tikz[baseline=0.05em] \fill [RestorationSub6] (0,0) rectangle (0.6em,0.6em); Outpainting (2.6\%) &
            \tikz[baseline=0.05em] \fill [RestorationSub7] (0,0) rectangle (0.6em,0.6em); Rain (2.5\%) \\[0.3em]

            \tikz[baseline=0.05em] \fill [RestorationSub8] (0,0) rectangle (0.6em,0.6em); Noise (2.2\%) &
            \tikz[baseline=0.05em] \fill [RestorationSub9] (0,0) rectangle (0.6em,0.6em); Ringing (2.2\%) &
            \tikz[baseline=0.05em] \fill [RestorationSub10] (0,0) rectangle (0.6em,0.6em); Watermark (2.1\%) &
            \tikz[baseline=0.05em] \fill [RestorationSub11] (0,0) rectangle (0.6em,0.6em); Others (15.0\%) \\[0.3em]
            
            \cellcolor{EnhancementMain}\textcolor{white}{Enhancement (21.9\%)} &
            \tikz[baseline=0.05em] \fill [EnhancementSub1] (0,0) rectangle (0.6em,0.6em); Contrast (3.6\%) &
            \tikz[baseline=0.05em] \fill [EnhancementSub2] (0,0) rectangle (0.6em,0.6em); Saturate (3.6\%) &
            \tikz[baseline=0.05em] \fill [EnhancementSub3] (0,0) rectangle (0.6em,0.6em); Color (3.2\%) \\[0.3em]

            \tikz[baseline=0.05em] \fill [EnhancementSub4] (0,0) rectangle (0.6em,0.6em); Brighten (1.8\%) &
            \tikz[baseline=0.05em] \fill [EnhancementSub5] (0,0) rectangle (0.6em,0.6em); Darken (1.8\%) &
            \tikz[baseline=0.05em] \fill [EnhancementSub6] (0,0) rectangle (0.6em,0.6em); Mosaic (1.8\%) &
            \tikz[baseline=0.05em] \fill [EnhancementSub7] (0,0) rectangle (0.6em,0.6em); Oversharp (1.8\%) \\[0.3em]

            \tikz[baseline=0.05em] \fill [EnhancementSub8] (0,0) rectangle (0.6em,0.6em); Retouch (0.5\%) &
            \tikz[baseline=0.05em] \fill [EnhancementSub9] (0,0) rectangle (0.6em,0.6em); LowLight (0.5\%) &
            \tikz[baseline=0.05em] \fill [EnhancementSub10] (0,0) rectangle (0.6em,0.6em); ColorCorr (0.5\%) &
            \tikz[baseline=0.05em] \fill [EnhancementSub11] (0,0) rectangle (0.6em,0.6em); Others (2.8\%) \\[0.3em]
            
            \cellcolor{AnnotationMain}\textcolor{white}{Dense Pred. (17.8\%)} &
            \tikz[baseline=0.05em] \fill [AnnotationSub1] (0,0) rectangle (0.6em,0.6em); Depth (5.6\%) &
            \tikz[baseline=0.05em] \fill [AnnotationSub2] (0,0) rectangle (0.6em,0.6em); Normal (3.1\%) &
            \tikz[baseline=0.05em] \fill [AnnotationSub3] (0,0) rectangle (0.6em,0.6em); HED (2.2\%) \\[0.3em]
            
            \tikz[baseline=0.05em] \fill [AnnotationSub4] (0,0) rectangle (0.6em,0.6em); Segmentation (1.8\%) &
            \tikz[baseline=0.05em] \fill [AnnotationSub5] (0,0) rectangle (0.6em,0.6em); Canny (1.8\%) &
            \tikz[baseline=0.05em] \fill [AnnotationSub6] (0,0) rectangle (0.6em,0.6em); Perc. Edge (1.8\%) &
            \tikz[baseline=0.05em] \fill [AnnotationMain!50] (0,0) rectangle (0.6em,0.6em); Others (1.5\%) \\[0.3em]
            
            \cellcolor{StylizationMain}\textcolor{white}{Stylization (10.6\%)} &
            \tikz[baseline=0.05em] \fill [StylizationSub1] (0,0) rectangle (0.6em,0.6em); Art Style (8.1\%) &
            \tikz[baseline=0.05em] \fill [StylizationSub2] (0,0) rectangle (0.6em,0.6em); LLF (0.5\%) &
            \tikz[baseline=0.05em] \fill [StylizationSub3] (0,0) rectangle (0.6em,0.6em); MultiTM (0.5\%) \\[0.3em]

            \tikz[baseline=0.05em] \fill [StylizationSub4] (0,0) rectangle (0.6em,0.6em); Pencil (0.5\%) &
            \tikz[baseline=0.05em] \fill [StylizationSub5] (0,0) rectangle (0.6em,0.6em); Photo. (0.5\%) &
            \tikz[baseline=0.05em] \fill [StylizationSub6] (0,0) rectangle (0.6em,0.6em); RTV (0.5\%) &
	        \end{tabular}%
	        }
	    \end{minipage}
	    \caption{OmniLV dataset distribution with main categories.}
	    \label{fig:dataset_composition}
\end{figure}

\subsection{Large-Scale OmniLV Dataset}
To build a universal low-level vision model, we construct a large-scale multi-task dataset containing 40 million instances over 100 sub-tasks across four major domains: image restoration, image enhancement, dense prediction, and stylization. The main categories and distribution of OmniLV dataset are illustrated in Fig.~\ref{fig:dataset_composition}. The dataset is sourced from publicly available collections and synthetically generated pairs, with additional high-quality data created through internal pipelines.

\noindent\textbf{Image Restoration.} The restoration dataset covers 23 major tasks with a total of 45 sub-tasks, addressing various degradation types such as motion blur, noise, and weather-induced distortions. It consists of both real-world degraded images and synthetic degradation pairs, carefully processed alongside high-quality ground truth images to ensure realism and diversity.

\noindent\textbf{Image Enhancement.} The enhancement dataset includes 14 major tasks with a total of 25 sub-tasks, covering tasks such as low-light correction, contrast enhancement, and saturation refinement. The dataset is composed of professionally edited reference images alongside algorithmically generated enhancement pairs, ensuring controlled transformations that align with perceptual quality.

\noindent\textbf{Weak-semantic Dense Prediction.} For dense prediction tasks, we compile annotated datasets for 10 tasks, including edge detection, depth estimation, and surface normal prediction. Each sample contains pixel-level ground truth annotations paired with descriptive task-specific instructions, facilitating multimodal learning.

\noindent\textbf{Image Stylization.} The stylization dataset spans 20 tasks, covering artistic transformations across various styles and techniques. It includes both real-world artistic works and style-transferred images generated by neural algorithms, ensuring a diverse range of stylish effects. We implement in-context learning on image stylization tasks due to the difficulty of defining task prompt. 

\noindent\textbf{Dataset Summary and Test Set Construction.}
OmniLV dataset comprises four major task categories with over 100 sub-tasks and approximately 40 million training instances. For publicly available datasets, we directly adopt their test sets for evaluation. For our synthesized tasks, we construct test sets based on DIV2K-val~\citep{agustsson2017ntire}, forming OmniLV-Test (OLV-T). OLV-T consists of 44 task-specific test sets, each containing 100 images, totaling 4,400 test images with 1K resolution. Further details on dataset partitioning and evaluation can be found in the supplementary material.

\subsection{Model Training and Sampling Settings}
\label{sec: training_setting}

The training of OmniLV is divided into three stages: In the first stage, we train the model with images at a resolution of $512^2$, focusing solely on single-image tasks. We use a constant learning rate of 1e-4 and train for 100k steps with a batch size of 512. The second stage adds in-context learning (ICL) tasks. We continue training for another 100k steps, maintaining the same learning rate of 1e-4 and batch size of 512. Finally, in the third stage, we increase the resolution to $1024^2$ and train on all tasks. The batch size is reduced to 128, and the learning rate remains at 1e-4. This final stage ensures that the model is trained to handle a variety of tasks and image sizes effectively. The model is trained using 16 A100 GPUs.

%% file: sec/4_experiment.tex
\section{Experiments}
\subsection{Comparisons with Existing Works}
\begin{table}[t]
    \centering
    \renewcommand{\arraystretch}{1.2}
    \caption{Quantitative comparison on restoration tasks. \textcolor{red}{Red} and \textcolor{blue}{blue} colors represent the best and second best performance, respectively, excluding specialized models. All values are reported as PSNR$\uparrow$ / MUSIQ$\uparrow$. For specialized models, if a model achieves the best value, the corresponding number is highlighted in \textbf{bold}.}
    \resizebox{\textwidth}{!}{%
    \footnotesize
    \begin{tabular}{l l *{10}{c}}
        \toprule
        & & \multicolumn{2}{c}{Deblur} & \multicolumn{1}{c}{Compression} & \multicolumn{2}{c}{Denoise} & \multicolumn{2}{c}{Derain} & \multicolumn{1}{c}{Desnow} & \multicolumn{1}{c}{BIR} & \multicolumn{1}{c}{Face} \\ 
        \cmidrule{3-12}
        \multirow{-2}{*}{Category} & \multirow{-2}{*}{Method} 
        & OLV-T(6 types) & RealBlur-J & OLV-T(2 types) & OLV-T(6 types) & SIDD & Synthetic & Rain1400 & Snow100K-L & DIV2K & CelebA \\ 
        \midrule
        
        \multirow{6}{*}{Specialized}
        & X-Restormer & 21.18/45.17 & 26.57/50.41 & -- & \textbf{27.19}/63.67 & 31.95/22.04 & 27.10/71.34 & 32.35/70.34 & -- & -- & -- \\
        & MPRNet      & 20.33/43.35 & 26.51/48.45 & -- & 24.53/45.66 & 39.63/22.34 & 25.16/69.40 & 32.04/69.98 & -- & -- & -- \\
        & MAXIM       & 21.39/44.13 & \textbf{29.99}/55.68 & -- & 24.75/49.30 & \textbf{39.68}/22.39 & 25.82/71.15 & 32.25/70.27 & -- & -- & -- \\
        & DiffBIR     & -- & -- & -- & -- & -- & -- & -- & -- & \textbf{22.77}/67.01 & -- \\
        & GFPGAN    & -- & -- & -- & -- & -- & -- & -- & -- & -- & \textbf{25.80}/69.76 \\ 
        & CodeFormer & -- & -- & -- & -- & -- & -- & -- & -- & -- & 25.15/75.55 \\
        & SUPIR & -- & 25.62/46.43 & -- & -- & 26.38/25.31 & -- & 21.25/69.29 & 17.82/62.73 & -- & 25.15/74.68 \\
        & DiT4SR & -- & 22.71/\textbf{66.17} & -- & -- & 26.96/\textbf{36.96} & -- & 18.66/\textbf{72.25} & 16.13/\textbf{69.47} & -- & 25.80/\textbf{76.36} \\
        \midrule
        
        \multirow{3}{*}{All-in-One Restoration}
        & X-Restormer & 21.44/39.73 & \textcolor{blue}{26.23}/38.84 & -- & \textcolor{blue}{25.96}/\textcolor{blue}{62.42} & 24.06/20.84 & \textcolor{blue}{23.28}/\textcolor{red}{69.00} & \textcolor{red}{32.12}/\textcolor{blue}{70.26} & -- & -- & -- \\
        & DA-CLIP   & 19.94/34.98 & 18.82/39.22 & -- & 22.99/44.89 & 26.40/29.25 & 23.15/53.18 & 26.44/67.78 & -- & -- & -- \\
        & AutoDIR  & 20.09/45.07 & 19.10/\textcolor{blue}{49.63} & -- & \textcolor{red}{26.46}/57.80 & 22.19/28.72 & \textcolor{red}{25.33}/{64.59} & \textcolor{blue}{26.21}/\textcolor{red}{70.75} & -- & -- & -- \\
        \midrule
        
        \multirow{3}{*}{Visual-Prompt-based}
        & Painter  & 17.05/28.74 & 15.37/28.79 & 17.84/34.43 & 18.04/37.11 & \textcolor{red}{38.65}/21.57 & 17.84/34.43 & 27.92/62.38 & 20.30/47.60 & -- & -- \\
        & PromptGIP  & 20.01/31.26 & 22.94/29.65 & 21.93/35.15 & 22.80/35.58 & 26.16/22.79 & 21.93/35.15 & 23.87/50.62 & 20.29/40.21 & -- & -- \\
        & GenLV   & \textcolor{blue}{22.15}/33.00 & 25.53/29.12 & \textcolor{red}{23.59}/35.96 & 23.51/38.21 & 30.41/\textcolor{red}{28.10} & 23.59/35.96 & 26.26/56.99 & 20.21/45.61 & -- & -- \\
        \midrule
        
        \multirow{2}{*}{Text-Prompt-based}
        & PromptFix & 20.32/43.75 & 26.14/39.37 & 18.10/54.01 & 14.59/51.77 & 24.25/21.22 & 18.10/54.01 & 21.61/63.07 & \textcolor{blue}{21.12}/53.83 & 13.77/29.49 & -- \\
        & Pixwizard & 17.90/\textcolor{blue}{64.19} & 23.34/\textcolor{red}{55.97} & 18.99/\textcolor{blue}{62.40}  & 17.22/63.05 & 27.60/\textcolor{blue}{23.63} & 18.99/62.40 & 23.84/66.89 & \textcolor{blue}{21.12}/\textcolor{red}{61.41} & \textcolor{blue}{19.03}/\textcolor{blue}{59.90} & -- \\ \midrule
        
        Multi-Modal Instruction & OmniLV & \textcolor{red}{22.57}/\textcolor{red}{68.95} & \textcolor{red}{28.24}/36.09 & \textcolor{blue}{22.93}/\textcolor{red}{68.99} & {23.53}/\textcolor{red}{69.23} & \textcolor{blue}{32.96}/22.42 & 22.93/\textcolor{blue}{68.99} & 24.98/65.66 & \textcolor{red}{24.57}/\textcolor{blue}{61.19} & \textcolor{red}{22.36}/\textcolor{red}{69.55} & \textcolor{red}{25.04}/\textcolor{red}{70.70} \\
        \bottomrule
    \end{tabular}%
    }
    \label{tab:restoration}
\end{table}

\begin{table}[t]
    \centering
    \renewcommand{\arraystretch}{1}
    \footnotesize
    \caption{Quantitative comparison on enhancement tasks.}
    \resizebox{\textwidth}{!}{%
    \begin{tabular}{l l ccccccc}
        \toprule
         & 
        & Brighten & Darken & Low light & Photoretouching & \multicolumn{1}{c}{Contrast Adjust} & {Saturation Adjust} & {Oversharpening} \\
        \cmidrule(lr){3-9}
        \multirow{-2}{*}{Category} & \multirow{-2}{*}{Method}
        & OLV-T(4 types) & OLV-T(4 types) & LOLv2-Real& MIT5K
        & OLV-T(4 types) 
        & OLV-T(4 types) & OLV-T(1 type) \\ 
        \midrule
        \multirow{3}{*}{Specialized} 
        & Retinexformer & -- & 16.72/65.73 & 22.79/59.30 & 16.12/63.24 & -- & -- & -- \\
        & MIRNet   & -- & 16.35/65.45 & 28.10/63.35 & 19.37/\textbf{65.59} & -- & -- & -- \\
        & MAXIM   & -- & 16.09/67.16 & \textbf{34.04/70.75} & 14.98/62.90 & -- & -- & -- \\
        \midrule
        \multirow{2}{*}{All-in-One Restoration} 
        & DA-CLIP   & -- & 14.91/53.04 & 26.64/\textcolor{blue}{67.73} & --         & -- & -- & -- \\
        & AutoDIR    & -- & 15.48/64.74 & 24.16/\textcolor{red}{67.91} & --         & -- & -- & -- \\
        \midrule
        \multirow{3}{*}{Visual-Prompt-based} 
       & Painter     & 12.00/36.77 & 13.96/37.09 & \textcolor{red}{29.44}/53.82 & 17.19/58.39 & 12.55/35.99 & 13.25/36.66 & -- \\
       & PromptGIP   & 15.46/35.43 & 17.85/33.89 & \textcolor{blue}{21.35}/38.18 & 16.57/43.02 & 15.80/33.75 & 16.63/34.49 & 16.63/38.74 \\
        & GenLV    & \textcolor{blue}{21.11}/40.16 & \textcolor{red}{21.70}/39.31 & 21.01/50.84 & \textcolor{red}{24.91}/56.08 & \textcolor{blue}{21.58}/39.46 & \textcolor{blue}{20.87}/40.29 & \textcolor{blue}{21.69}/37.08 \\

        \midrule
        \multirow{2}{*}{Text-Prompt-based} 
        & PromptFix  & 10.55/57.78 & 10.15/54.40 & 17.16/63.54 & 11.09/52.89 & 11.34/57.76 & 12.31/58.49 & 14.93/56.01 \\
        & PixWizard   & 11.16/\textcolor{blue}{64.14} & 13.81/\textcolor{blue}{65.44} & 14.07/62.11 & 15.99/\textcolor{red}{63.59} & 13.12/\textcolor{blue}{65.51} & 12.77/\textcolor{blue}{65.13} & 13.55/\textcolor{blue}{71.00} \\
        \midrule
        Multi-Modal Instruction & OmniLV        & \textcolor{red}{22.58}/\textcolor{red}{70.54} & \textcolor{blue}{20.28}/\textcolor{red}{69.77} & 18.60/58.76 & \textcolor{blue}{19.78}/\textcolor{blue}{62.12} & \textcolor{red}{20.91}/\textcolor{red}{69.95} & \textcolor{red}{21.80}/\textcolor{red}{70.91} & \textcolor{red}{23.64}/\textcolor{red}{70.87} \\
        \bottomrule
    \end{tabular}%
    }
    \label{tab:enhancement}
\end{table}

\begin{table}[t]
    \centering
    \caption{Quantitative comparison on stylization tasks and weak-semantic dense prediction tasks.}
    \label{tab:comparison_style_dense}
    \renewcommand{\arraystretch}{1.00}  
    \scriptsize  
    \setlength{\tabcolsep}{4pt}  
    
    \begin{minipage}[b]{0.47\textwidth}
        \vspace{0pt} 
        \begin{subtable}{\textwidth}
            \centering
            \caption{Stylization Tasks.}
            \resizebox{\textwidth}{!}{%
            \begin{tabular}{l l cc cc}
                \toprule
                & &\multicolumn{2}{c}{LLF} & \multicolumn{2}{c}{PencilDrawing} \\
               \cmidrule{3-6}
               \multirow{-2}{*}{Category} & \multirow{-2}{*}{Method} 
               & PSNR$\uparrow$ & FID$\downarrow$  & PSNR$\uparrow$ & FID$\downarrow$\\
               \midrule
               \multirow{3}{*}{Visual-Prompt-based}                  & Painter & 13.64 & 120.3 & 8.434 & 157.5 \\
               & PromptGIP & \textcolor{blue}{22.87} & 50.61 & \textcolor{blue}{21.35} & 132.5 \\
               & GenLV & \textcolor{red}{25.66} & \textcolor{blue}{29.53} & \textcolor{red}{28.29} & \textcolor{red}{38.70} \\
               \midrule
               Multi-Modal Instruction & OmniLV & 21.72 & \textcolor{red}{24.16} & 20.33 & \textcolor{blue}{54.18} \\
               \bottomrule
            \end{tabular}
            }
            \label{tab:style}
        \end{subtable}
    \end{minipage}%
    \hfill 
    \vspace{1em}
    \begin{minipage}[b]{0.5\textwidth}
        \vspace{0pt} 
        \begin{subtable}{\textwidth}
            \centering
             \caption{Dense Prediction Tasks. }
             \label{tab:dense}
            \resizebox{\textwidth}{!}{%
            \begin{tabular}{l c c c c c c}
                \toprule
                & \multicolumn{1}{c}{Depth Esti.} & \multicolumn{1}{c}{Normal Esti.} & \multicolumn{1}{c}{HED} & \multicolumn{1}{c}{DA-2K} & \multicolumn{1}{c}{SUNRGB-D} & \multicolumn{1}{c}{PASCAL-Ctx} \\
                \cmidrule{2-7}
                Method & RMSE$\downarrow$ & Mean Angle Error$\downarrow$ & MAE$\downarrow$ & Accuracy$\uparrow$ & RMSE$\downarrow$ & Score$\downarrow$ \\
                \midrule
                PromptGIP & -- & -- & 85.90 & -- & -- & -- \\
                GenLV & -- & -- & 44.28 & -- & -- & -- \\
                D.A. & \textcolor{red}{0.291} & -- & -- & -- & -- & -- \\
                InvPT & -- & \textcolor{blue}{19.04} & -- & -- & -- & -- \\
                PixWizard & 0.941 & 19.65 & \textcolor{blue}{25.95} & \textcolor{blue}{85.49} & \textcolor{blue}{0.3523} & \textcolor{blue}{34.65} \\
                \midrule
                OmniLV & \textcolor{blue}{0.525} & \textcolor{red}{17.30} & \textcolor{red}{20.75} & \textcolor{red}{90.86} & \textcolor{red}{0.3018} & \textcolor{red}{29.91} \\
                \bottomrule
            \end{tabular}
            }
        \end{subtable}
    \end{minipage}
    \raggedright\footnotesize
\end{table}

\begin{figure}[t]
    \centering
    \includegraphics[width=\textwidth]{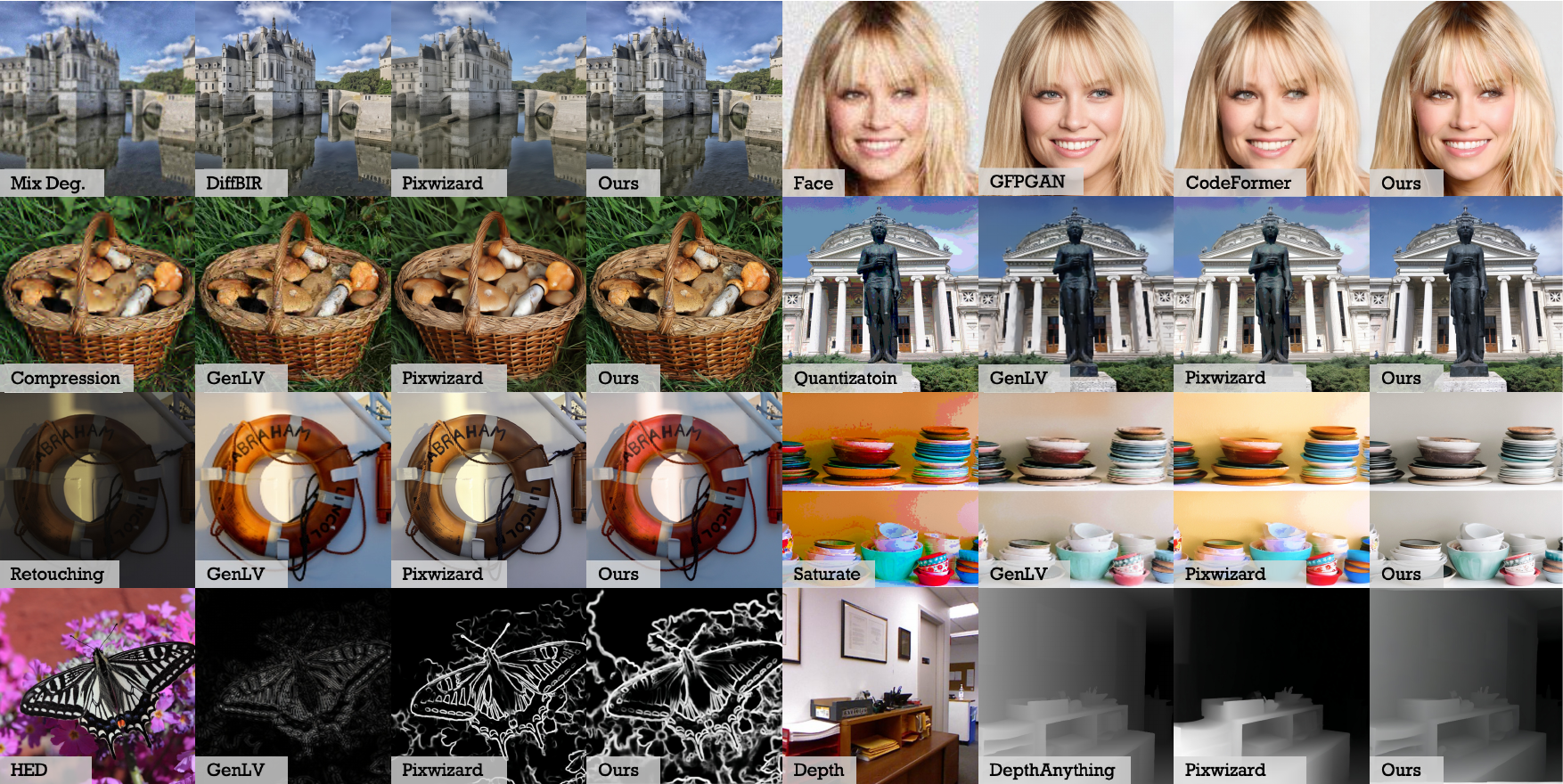}
    \caption{Comparison results for low-level vision tasks. More results can be found in the supplementary material.}
    \label{fig:comparison_main}
    \vspace{-2em}
\end{figure}

As a universal model, OmniLV exhibits strong performance across diverse low-level vision tasks, even compared with existing task-specific models. We compare OmniLV against four groups of baselines: task-specific methods (X-Restormer \citep{chen2024comparative}, MPRNet \citep{Zamir2021MPRNet}, MAXIM \citep{tu2022maxim}, DiffBIR \citep{lin2024diffbir}, GFPGAN \citep{wang2021gfpgan}, CodeFormer \citep{zhou2022codeformer}, Retinexformer \citep{cai2023retinexformer}, MIRNet \citep{Zamir2020MIRNet}, SUPIR~\citep{yu2024scaling}, DiT4SR~\citep{duan2025dit4sr} and Depth Anything (D.A.) \citep{yang2024depth}); all-in-one methods (DA-CLIP \citep{luo2023controlling} and AutoDIR \citep{jiang2023autodir}); visual-prompt methods (PromptGIP \citep{liu2023unifying}, GenLV \citep{chen2024learning}, and Painter \citep{Painter}); and text-guided diffusion methods (PixWizard \citep{lin2024pixwizard} and PromptFix \citep{yu2024promptfix}).
Some of them are constrained to generating images of fixed size. In our comparison, we keep each method's original inference configuration and resize the generated image to target image size to facilitate fair comparisons.
We selected full-reference metric PSNR and no-reference metric MUSIQ~\citep{ke2021musiq} for quantitative comparison. 
The test sets include both our synthetic OLV-T and public datasets (RealBlur-J~\citep{rim_2020_ECCV}, SIDD~\citep{abdelhamed2018high}, Rain1400~\citep{fu2017removing}, Snow100K-L~\citep{liu2018desnownet}, DIV2K~\citep{agustsson2017ntire}, CelebA~\citep{liu2018large}, LOL-v2-Real~\citep{lolV2}, MIT5K~\citep{mit5k}). The supplementary material provides detailed results, evaluation protocol, inference cost, and FLOPs accounting.

\noindent\textbf{Image Restoration.} Tab.~\ref{tab:restoration} demonstrates that OmniLV achieves the highest PSNR scores across all restoration benchmarks when compared with diffusion-based models. We demonstrate the qualitative comparisons in Fig.~\ref{fig:comparison_main} on general low-level tasks. In addition, the MUSIQ scores of OmniLV are highly competitive on most benchmarks, further underscoring its strong performance. Notably, on the Blind Image Restoration (BIR) and Face benchmarks, OmniLV, as a universal model, attains performance levels that are comparable to those of state-of-the-art specialized models, thereby validating its effectiveness in handling diverse restoration tasks.

\noindent\textbf{Image Enhancement.} As reported in Tab.~\ref{tab:enhancement}, OmniLV significantly improves upon existing enhancement methods. The improvements highlight OmniLV’s ability to effectively enhance image quality while maintaining natural details and color fidelity.

\noindent\textbf{Dense Prediction.} Tab.~\ref{tab:dense} reports depth estimation (RMSE), normal estimation (mean angle error), edge detection (HED MAE), and additional dense benchmarks (DA-2K~\citep{yang2024depth}, SUNRGB-D~\citep{zhou2014learning}, and PASCAL-Context~\citep{mottaghi_cvpr14}). OmniLV improves over generalist/prompt-based baselines, while specialized models still lead on some single-task metrics (e.g., Depth Anything on depth with RMSE $0.291$). This indicates both the feasibility of unified dense prediction and remaining headroom versus task-specific systems; evaluation details are provided in the supplementary material.

\noindent\textbf{Stylization.} Tab.~\ref{tab:style} illustrates that OmniLV also performs well on stylization tasks such as Local Laplacian Filter (LLF) and Pencil Drawing~\citep{lu2012combining}. Since stylization tasks are challenging to describe using natural language, we employed visual prompts to guide the model in processing images. OmniLV obtains balanced results in terms of objective quality and perceptual quality, thus validating its versatility across diverse low-level vision tasks.

\subsection{More Exploration}
\label{sec:exploration}

\noindent\textbf{Text Prompt vs. Visual Prompt.} Our model supports both text prompt and visual prompt to guide the generation process. In Tab.~\ref{tab:text_visual}, we present a detailed comparison between the two prompting methods across several low-level vision tasks, including deblurring, denoising, contrast adjustment, and saturation adjustment. Adopting both prompts can yield better quality scores.



\noindent\textbf{Relationship to High-Semantic Tasks.} We further investigate the relationship between low-level vision tasks and high-semantic tasks such as image generation or image editing~\citep{sheynin2024emu, shi2020benchmark,yildirim2023instinpaint,hui2024hq,brooks2022instructpix2pix,Zhang2023MagicBrush,zhao2024ultraeditinstructionbasedfinegrainedimage}. As shown in Tab.~\ref{tab:relationship}, when high-semantic tasks are included in the training data, performance on low-level vision tasks degrades. Specifically, performance for various tasks is consistently lower when high-semantic tasks are incorporated. We hypothesize this occurs because high-semantic tasks prioritize conceptual coherence and structural abstraction over pixel-accurate reconstruction, which conflicts with objectives that demand fine-grained texture recovery and precise detail preservation.


\newcommand{\twopanelH}{3.6cm}

\begin{figure*}[t]
\centering
\begin{minipage}[t]{0.40\textwidth}
  \centering
  \includegraphics[height=\twopanelH,keepaspectratio]{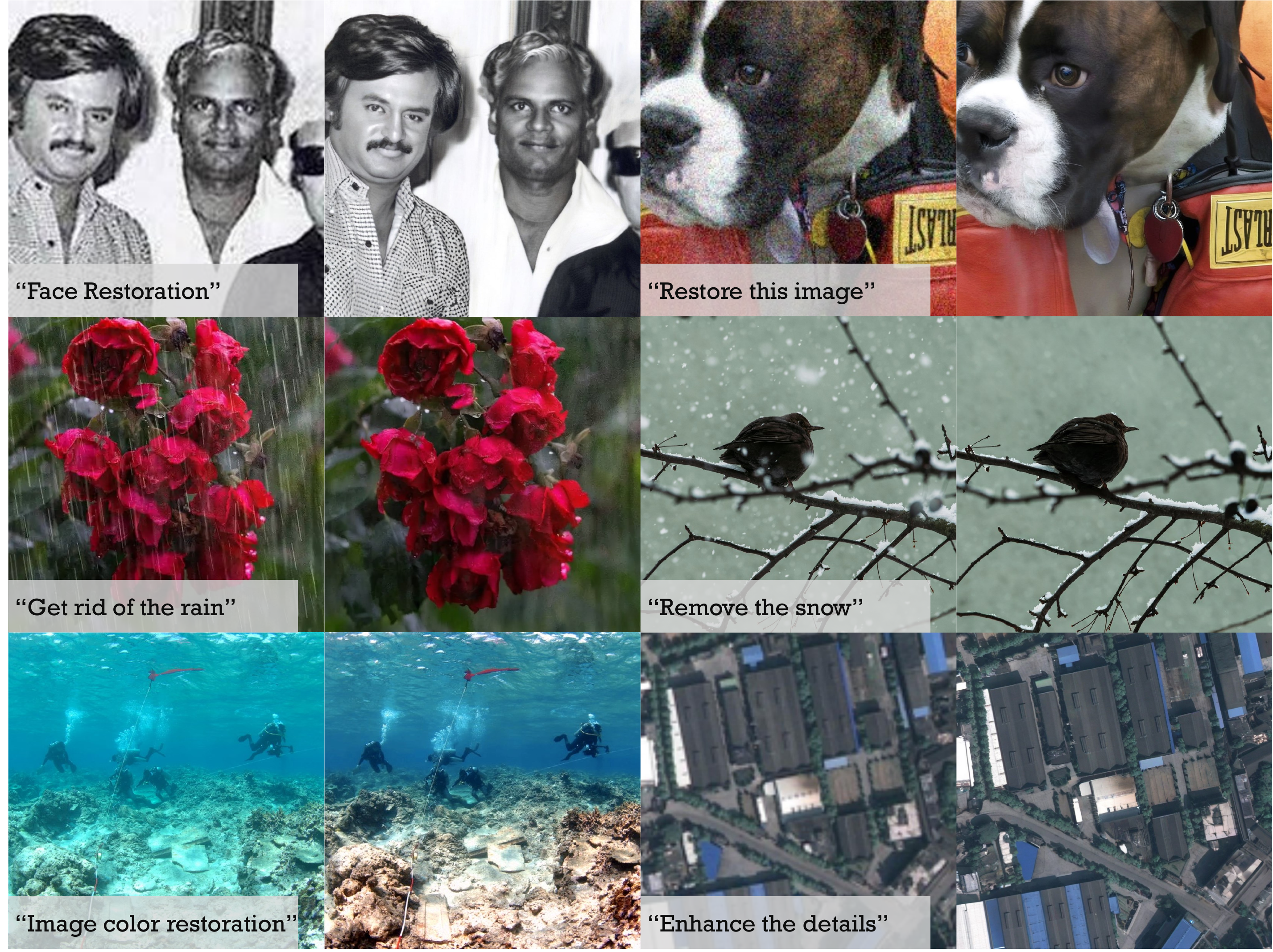}
\end{minipage}\hfill
\begin{minipage}[t]{0.6\textwidth}
  \centering
  \includegraphics[height=\twopanelH,keepaspectratio]
  {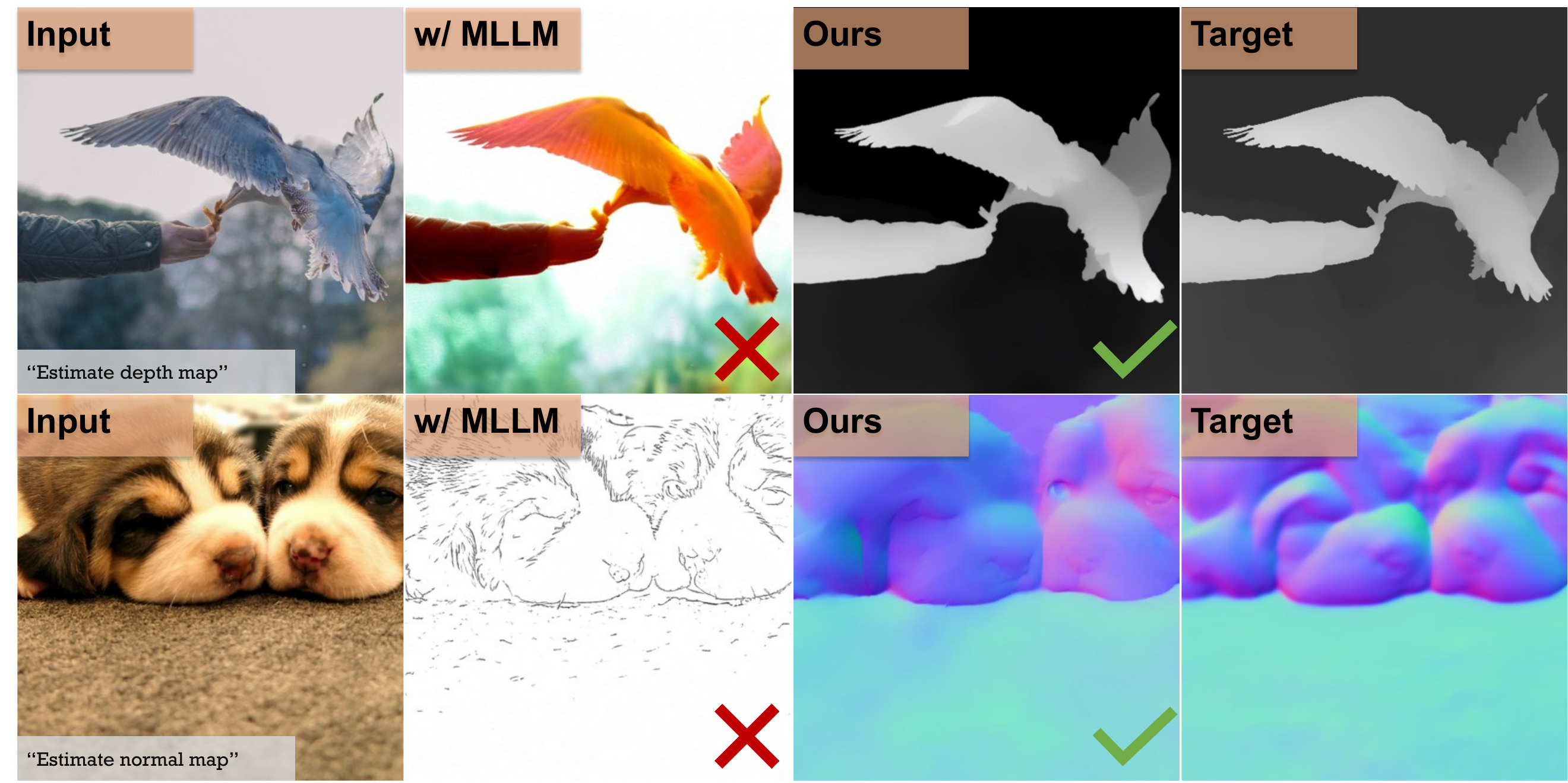}
\end{minipage}
\caption{(Left)Examples of image restoration in various scenarios. (Right)Task mismatch samples.}
\label{fig:task_mismatch_sample_real_data}
\vspace{-2em}
\end{figure*}

\noindent\textbf{Generalization Exploration.} We investigated the generalizability of OmniLV in terms of domain-specific adaptation and real-world robustness. Specifically, we selected images with various real-world degradations to comprehensively assess the model's performance. As shown in Fig.~\ref{fig:task_mismatch_sample_real_data}, OmniLV effectively restores images in these diverse conditions, demonstrating its robustness and versatility in handling complex real-world degradations, such as real-world restoration, deraining, desnowing, underwater image enhancement, and satellite image enhancement.

\begin{figure}[!ht]
\vspace{-1em}
    \centering
    \includegraphics[width=\linewidth]{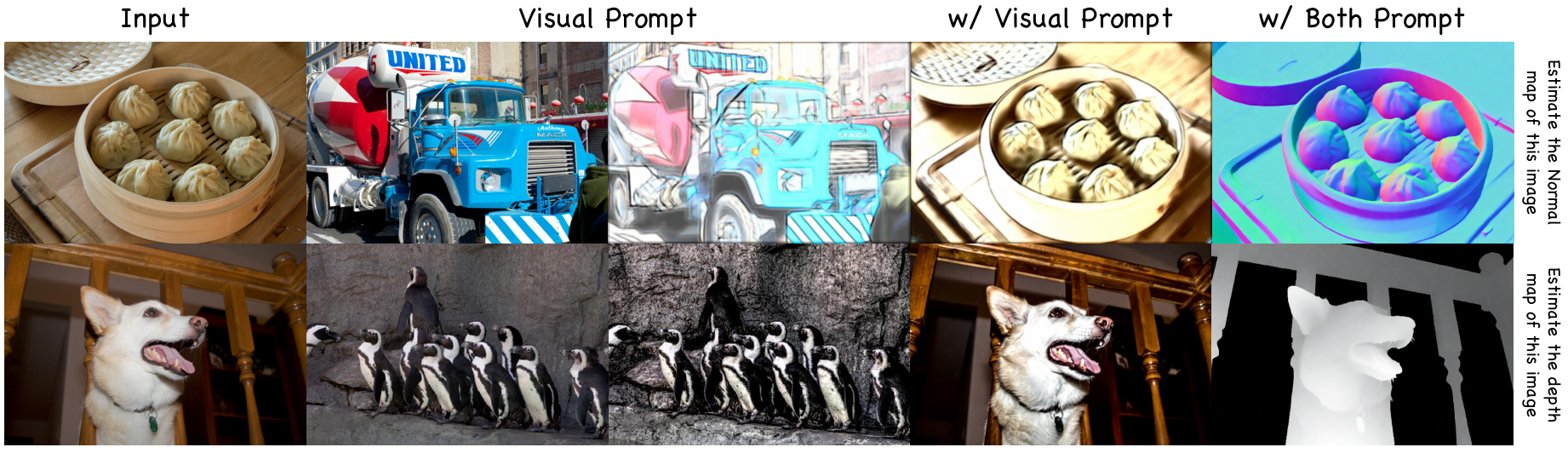}
    \caption{Results when providing conflict visual prompt and text instruction. Our model tends to follow text instruction when conflict happens.}
    \label{fig:prompt_conflict}
\vspace{-1em}
\end{figure}

\noindent\textbf{Prompt Conflict.}
To study how OmniLV behaves under inconsistent conditioning, we construct deliberately conflicting cases where the text instruction and the visual exemplars specify different tasks. As illustrated in Fig.~\ref{fig:prompt_conflict}, when both prompts are provided, the output consistently follows the text instruction rather than the visual exemplars. We do not intentionally create such conflicting pairs in our training data. We hypothesize this occurs because the training distribution contains more text-specified instructions than conflicting visual exemplars, biasing the model toward text in ambiguous cases. Consequently, the model has learned to treat the text instruction as the primary source of task specification in the presence of conflicts.

\begin{figure}[!ht]
    \centering
    \includegraphics[width=\linewidth]{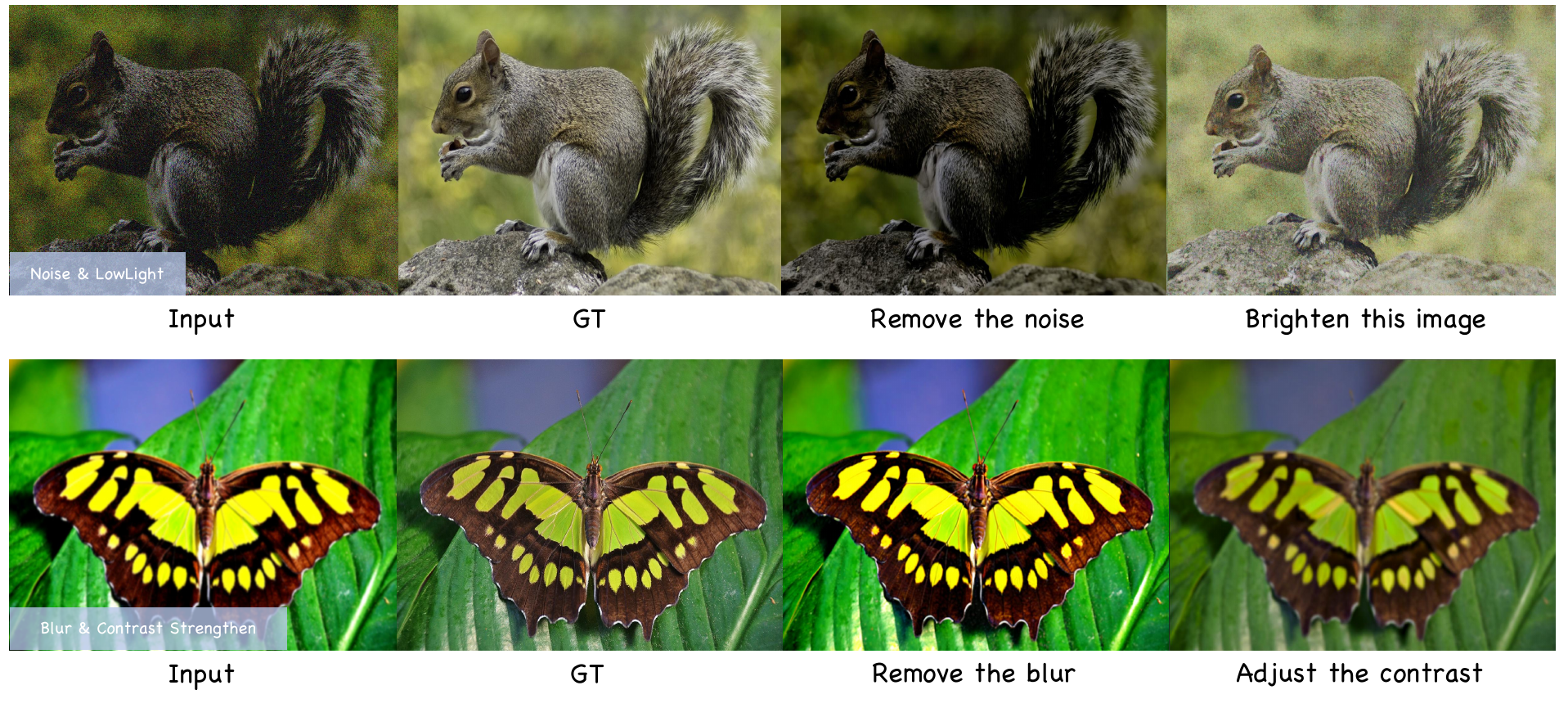}
    \caption{Results for cross-domain tasks. For cross-domain tasks, OmniLV can generate correct results with regard to the prompts.}
    \label{fig:crossdomain}
\end{figure}

\noindent\textbf{Cross-domain Task.}
Following the definition in GenLV~\citep{chen2024learning}, we categorize low-level vision tasks into five domains: restoration, enhancement, stylization, dense prediction, and natural images. The cross-domain capability in our setting refers to a model’s ability to not only accomplish tasks within a single domain but also handle tasks from multiple domains simultaneously. To assess this ability, we construct composite tasks that involve multiple domains at once. As illustrated in Fig.~\ref{fig:crossdomain}, OmniLV demonstrates strong instruction-following behavior and consistently produces correct results across these mixed-domain scenarios.

\noindent\textbf{Instruction Following Ability.}
\label{Instruction Following Ability}
Existing models often fail to adhere to the instructions and perform wrong transformations, especially for dense prediction tasks. In contrast, OmniLV shows strong instruction-following ability courtesy of the dedicated design choice described in Sec.~\ref{sec:multimodel}. To illustrate it more clearly, we compute the success rates of OmniLV on some dense prediction tasks compared to OmniGen~\citep{xiao2024omnigen} and PixWizard~\citep{lin2024pixwizard}. We deem a case as successful when the model performs the intended task, no matter how well. Tab.~\ref{tab:instruction_following} and Fig.~\ref{fig:task_mismatch_sample_real_data} presents the results, showing the superior task alignment ability of OmniLV.

\begin{table}[!t]
    \centering
    \begin{minipage}[t]{0.60\linewidth}
        \centering
        \captionof{table}{Successful rates on dense prediction tasks.}
        \vspace{1.2em}
        \label{tab:instruction_following}
        \small
        \begin{tabular}{@{}cccc@{}}
                \toprule
                Models  & Image2Canny & Image2Depth & Image2HED \\
                \midrule
                OmniGen & 80\% & 75\% & - \\
                Pixwizard & 100\% & 93\% & 97\% \\
                OmniLV & 100\% & 100\% & 100\% \\
                \bottomrule
            \end{tabular}
    \end{minipage}
    \hfill
    \begin{minipage}[t]{0.36\linewidth}
        \centering
        \captionof{table}{Human-preference Elo ratings on benchmarks.}
        \label{tab:elo_omnilv}
        \small
        \begin{tabular}{lc}
                \toprule
                \textbf{Model} & \textbf{Elo rating} \\
                \midrule
                OmniLV & 1,475 \\
                PixWizard   & 1,310 \\
                PromptFix   & 1,000 \\
                \bottomrule
            \end{tabular}
    \end{minipage}
\end{table}

\noindent\textbf{Human Preference.}
We conduct pairwise human preference comparisons for evaluating low-level image restoration and enhancement quality. As summarized in Tab.~\ref{tab:elo_omnilv}, OmniLV achieves the highest human-preference Elo score, outperforming both PixWizard and PromptFix by substantial margins. The supplementary material provides the full study protocol and Elo aggregation details.

\section{Discussion \& Limitation}
\vspace{-1em}
OmniLV targets a unified low-level vision generalist that supports diverse low-level vision tasks with a single set of weights. Our contribution lies in the unification setting and the empirically validated design choices that improve task alignment across domains, rather than proposing a new diffusion backbone architecture. However, OmniLV does not always achieve optimal performance in certain specialized scenarios. Future work will focus on enhancing task-specific performance by refining model components and training strategies. 

%% file: sec/5_conclusion.tex
\vspace{-0.5em}
\section{Conclusion}
\vspace{-0.5em}
In this work, we introduced \textbf{OmniLV}, a unified multimodal framework for low-level vision that successfully handles over 100 sub-tasks, including image restoration, enhancement, weak-semantic dense prediction, and stylization. By leveraging both textual and visual prompts with generative priors, OmniLV demonstrates robust generalization, high-fidelity results, and flexibility across arbitrary resolutions. OmniLV achieves state-of-the-art performance in multiple low-level vision tasks and demonstrates promising generalization capabilities in real-world scenarios.


\vspace{0.5em}
\noindent\textbf{Acknowledgements.} Supported by Shanghai Artificial Intelligence Laboratory.

\vspace{1em}

%% file: main.bib
@String(CVPR= {IEEE Conf. Comput. Vis. Pattern Recog.})

@String(ECCV= {Eur. Conf. Comput. Vis.})

@String(AAAI = {AAAI})

@String(CVPR  = {CVPR})

@String(ECCV  = {ECCV})

@inproceedings{duan2025dit4sr,
  title={Dit4sr: Taming diffusion transformer for real-world image super-resolution},
  author={Duan, Zheng-Peng and Zhang, Jiawei and Jin, Xin and Zhang, Ziheng and Xiong, Zheng and Zou, Dongqing and Ren, Jimmy S and Guo, Chunle and Li, Chongyi},
  booktitle={Proceedings of the IEEE/CVF International Conference on Computer Vision},
  pages={18948--18958},
  year={2025}
}

@InProceedings{mottaghi_cvpr14,
 author       = {Roozbeh Mottaghi and Xianjie Chen and Xiaobai Liu and Nam-Gyu Cho and Seong-Whan Lee and Sanja Fidler and Raquel Urtasun and Alan Yuille},
 title        = {The Role of Context for Object Detection and Semantic Segmentation in the Wild},
 booktitle    = {IEEE Conference on Computer Vision and Pattern Recognition (CVPR)},
 year         = {2014},
}

@article{zhou2014learning,
  title={Learning deep features for scene recognition using places database},
  author={Zhou, Bolei and Lapedriza, Agata and Xiao, Jianxiong and Torralba, Antonio and Oliva, Aude},
  journal={Advances in neural information processing systems},
  volume={27},
  year={2014}
}

@article{deng2025emerging,
  title={Emerging properties in unified multimodal pretraining},
  author={Deng, Chaorui and Zhu, Deyao and Li, Kunchang and Gou, Chenhui and Li, Feng and Wang, Zeyu and Zhong, Shu and Yu, Weihao and Nie, Xiaonan and Song, Ziang and others},
  journal={arXiv preprint arXiv:2505.14683},
  year={2025}
}

@article{wu2025omnigen2,
  title={Omnigen2: Exploration to advanced multimodal generation},
  author={Wu, Chenyuan and Zheng, Pengfei and Yan, Ruiran and Xiao, Shitao and Luo, Xin and Wang, Yueze and Li, Wanli and Jiang, Xiyan and Liu, Yexin and Zhou, Junjie and others},
  journal={arXiv preprint arXiv:2506.18871},
  year={2025}
}

@inproceedings{yang2024pixel,
  title={Pixel-aware stable diffusion for realistic image super-resolution and personalized stylization},
  author={Yang, Tao and Wu, Rongyuan and Ren, Peiran and Xie, Xuansong and Zhang, Lei},
  booktitle={European Conference on Computer Vision},
  pages={74--91},
  year={2024},
  organization={Springer}
}

@article{wang2024exploiting,
  title={Exploiting diffusion prior for real-world image super-resolution},
  author={Wang, Jianyi and Yue, Zongsheng and Zhou, Shangchen and Chan, Kelvin CK and Loy, Chen Change},
  journal={International Journal of Computer Vision},
  volume={132},
  number={12},
  pages={5929--5949},
  year={2024},
  publisher={Springer}
}

@inproceedings{lin2024diffbir,
  title={Diffbir: Toward blind image restoration with generative diffusion prior},
  author={Lin, Xinqi and He, Jingwen and Chen, Ziyan and Lyu, Zhaoyang and Dai, Bo and Yu, Fanghua and Qiao, Yu and Ouyang, Wanli and Dong, Chao},
  booktitle={European Conference on Computer Vision},
  pages={430--448},
  year={2024},
  organization={Springer}
}

@inproceedings{yu2024scaling,
  title={Scaling up to excellence: Practicing model scaling for photo-realistic image restoration in the wild},
  author={Yu, Fanghua and Gu, Jinjin and Li, Zheyuan and Hu, Jinfan and Kong, Xiangtao and Wang, Xintao and He, Jingwen and Qiao, Yu and Dong, Chao},
  booktitle={Proceedings of the IEEE/CVF Conference on Computer Vision and Pattern Recognition},
  pages={25669--25680},
  year={2024}
}

@inproceedings{fu2017removing,
  title={Removing rain from single images via a deep detail network},
  author={Fu, Xueyang and Huang, Jiabin and Zeng, Delu and Huang, Yue and Ding, Xinghao and Paisley, John},
  booktitle={Proceedings of the IEEE conference on computer vision and pattern recognition},
  pages={3855--3863},
  year={2017}
}

@inproceedings{peebles2023scalable,
  title={Scalable diffusion models with transformers},
  author={Peebles, William and Xie, Saining},
  booktitle={Proceedings of the IEEE/CVF international conference on computer vision},
  pages={4195--4205},
  year={2023}
}

@inproceedings{lu2012combining,
  title={Combining sketch and tone for pencil drawing production},
  author={Lu, Cewu and Xu, Li and Jia, Jiaya},
  booktitle={Proceedings of the symposium on non-photorealistic animation and rendering},
  pages={65--73},
  year={2012}
}

@article{liu2018desnownet,
  title={Desnownet: Context-aware deep network for snow removal},
  author={Liu, Yun-Fu and Jaw, Da-Wei and Huang, Shih-Chia and Hwang, Jenq-Neng},
  journal={IEEE Transactions on Image Processing},
  volume={27},
  number={6},
  pages={3064--3073},
  year={2018},
  publisher={IEEE}
}

@inproceedings{abdelhamed2018high,
  title={A high-quality denoising dataset for smartphone cameras},
  author={Abdelhamed, Abdelrahman and Lin, Stephen and Brown, Michael S},
  booktitle={Proceedings of the IEEE conference on computer vision and pattern recognition},
  pages={1692--1700},
  year={2018}
}

@inproceedings{agustsson2017ntire,
  title={Ntire 2017 challenge on single image super-resolution: Dataset and study},
  author={Agustsson, Eirikur and Timofte, Radu},
  booktitle={Proceedings of the IEEE conference on computer vision and pattern recognition workshops},
  pages={126--135},
  year={2017}
}

@misc{zhao2024ultraeditinstructionbasedfinegrainedimage,
      title={UltraEdit: Instruction-based Fine-Grained Image Editing at Scale}, 
      author={Haozhe Zhao and Xiaojian Ma and Liang Chen and Shuzheng Si and Rujie Wu and Kaikai An and Peiyu Yu and Minjia Zhang and Qing Li and Baobao Chang},
      year={2024},
      eprint={2407.05282},
      archivePrefix={arXiv},
      primaryClass={cs.CV},
      url={https://arxiv.org/abs/2407.05282}, 
}

@inproceedings{Zhang2023MagicBrush,
      title={MagicBrush: A Manually Annotated Dataset for Instruction-Guided Image Editing},
      author={Kai Zhang and Lingbo Mo and Wenhu Chen and Huan Sun and Yu Su},
      booktitle={Advances in Neural Information Processing Systems},
      year={2023}
    }

@article{brooks2022instructpix2pix,
  title={InstructPix2Pix: Learning to Follow Image Editing Instructions},
  author={Brooks, Tim and Holynski, Aleksander and Efros, Alexei A},
  journal={arXiv preprint arXiv:2211.09800},
  year={2022}
}

@article{hui2024hq,
  title   = {HQ-Edit: A High-Quality Dataset for Instruction-based Image Editing},
  author  = {Hui, Mude and Yang, Siwei and Zhao, Bingchen and Shi, Yichun and Wang, Heng and Wang, Peng and Zhou, Yuyin and Xie, Cihang},
  journal = {arXiv preprint arXiv:2404.09990},
  year    = {2024}
}

@misc{yildirim2023instinpaint,
      title={Inst-Inpaint: Instructing to Remove Objects with Diffusion Models}, 
      author={Ahmet Burak Yildirim and Vedat Baday and Erkut Erdem and Aykut Erdem and Aysegul Dundar},
      year={2023},
      eprint={2304.03246},
      archivePrefix={arXiv},
      primaryClass={cs.CV}
}

@inproceedings{shi2020benchmark,
  title={A Benchmark and Baseline for Language-Driven Image Editing},
  author={Shi, Jing and Xu, Ning and Bui, Trung and Dernoncourt, Franck and Wen, Zheng and Xu, Chenliang},
  booktitle={Proceedings of the Asian Conference on Computer Vision},
  year={2020}
}

@inproceedings{sheynin2024emu,
  title={Emu edit: Precise image editing via recognition and generation tasks},
  author={Sheynin, Shelly and Polyak, Adam and Singer, Uriel and Kirstain, Yuval and Zohar, Amit and Ashual, Oron and Parikh, Devi and Taigman, Yaniv},
  booktitle={Proceedings of the IEEE/CVF Conference on Computer Vision and Pattern Recognition},
  pages={8871--8879},
  year={2024}
}

@article{liu2018large,
  title={Large-scale celebfaces attributes (celeba) dataset},
  author={Liu, Ziwei and Luo, Ping and Wang, Xiaogang and Tang, Xiaoou},
  journal={Retrieved August},
  volume={15},
  number={2018},
  pages={11},
  year={2018}
}

@inproceedings{rim_2020_ECCV,
 title={Real-World Blur Dataset for Learning and Benchmarking Deblurring Algorithms},
 author={Rim, Jaesung and Lee, Haeyun and Won, Jucheol and Cho, Sunghyun},
 booktitle={Proceedings of the European Conference on Computer Vision (ECCV)},
 year={2020}
}

@article{yue2024arbitrary,
  title={Arbitrary-steps Image Super-resolution via Diffusion Inversion},
  author={Yue, Zongsheng and Liao, Kang and Loy, Chen Change},
  journal={arXiv preprint arXiv:2412.09013},
  year={2024}
}

@article{chen2024adversarial,
  title={Adversarial diffusion compression for real-world image super-resolution},
  author={Chen, Bin and Li, Gehui and Wu, Rongyuan and Zhang, Xindong and Chen, Jie and Zhang, Jian and Zhang, Lei},
  journal={arXiv preprint arXiv:2411.13383},
  year={2024}
}

@article{wu2024one,
  title={One-step effective diffusion network for real-world image super-resolution},
  author={Wu, Rongyuan and Sun, Lingchen and Ma, Zhiyuan and Zhang, Lei},
  journal={Advances in Neural Information Processing Systems},
  volume={37},
  pages={92529--92553},
  year={2024}
}

@inproceedings{ai2024dreamclear,
    title={DreamClear: High-Capacity Real-World Image Restoration with Privacy-Safe Dataset Curation},
    author={Yuang Ai and Xiaoqiang Zhou and Huaibo Huang and Xiaotian Han and Zhengyu Chen and Quanzeng You and Hongxia Yang},
    booktitle={The Thirty-eighth Annual Conference on Neural Information Processing Systems},
    year={2024},
    url={https://openreview.net/forum?id=6eoGVqMiIj}
}

@inproceedings{wang2019underexposed,
  title={Underexposed photo enhancement using deep illumination estimation},
  author={Wang, Ruixing and Zhang, Qing and Fu, Chi-Wing and Shen, Xiaoyong and Zheng, Wei-Shi and Jia, Jiaya},
  booktitle={Proceedings of the IEEE/CVF conference on computer vision and pattern recognition},
  pages={6849--6857},
  year={2019}
}

@inproceedings{chen2021hdrunet,
  title={Hdrunet: Single image hdr reconstruction with denoising and dequantization},
  author={Chen, Xiangyu and Liu, Yihao and Zhang, Zhengwen and Qiao, Yu and Dong, Chao},
  booktitle={Proceedings of the IEEE/CVF Conference on Computer Vision and Pattern Recognition},
  pages={354--363},
  year={2021}
}

@inproceedings{chen2021new,
  title={A new journey from SDRTV to HDRTV},
  author={Chen, Xiangyu and Zhang, Zhengwen and Ren, Jimmy S and Tian, Lynhoo and Qiao, Yu and Dong, Chao},
  booktitle={Proceedings of the IEEE/CVF International Conference on Computer Vision},
  pages={4500--4509},
  year={2021}
}

@inproceedings{ke2021musiq,
  title={Musiq: Multi-scale image quality transformer},
  author={Ke, Junjie and Wang, Qifei and Wang, Yilin and Milanfar, Peyman and Yang, Feng},
  booktitle={Proceedings of the IEEE/CVF international conference on computer vision},
  pages={5148--5157},
  year={2021}
}

@inproceedings{Zamir2020MIRNet,
    title={Learning Enriched Features for Real Image Restoration and Enhancement},
    author={Syed Waqas Zamir and Aditya Arora and Salman Khan and Munawar Hayat
            and Fahad Shahbaz Khan and Ming-Hsuan Yang and Ling Shao},
    booktitle={ECCV},
    year={2020}
}

@inproceedings{cai2023retinexformer,
  title={Retinexformer: One-stage retinex-based transformer for low-light image enhancement},
  author={Cai, Yuanhao and Bian, Hao and Lin, Jing and Wang, Haoqian and Timofte, Radu and Zhang, Yulun},
  booktitle={Proceedings of the IEEE/CVF international conference on computer vision},
  pages={12504--12513},
  year={2023}
}

@article{lin2024pixwizard,
  title={PixWizard: Versatile Image-to-Image Visual Assistant with Open-Language Instructions},
  author={Lin, Weifeng and Wei, Xinyu and Zhang, Renrui and Zhuo, Le and Zhao, Shitian and Huang, Siyuan and Xie, Junlin and Qiao, Yu and Gao, Peng and Li, Hongsheng},
  journal={arXiv preprint arXiv:2409.15278},
  year={2024}
}

@inproceedings{yu2024promptfix,
  title={PromptFix: You Prompt and We Fix the Photo},
  author={Yu, Yongsheng and Zeng, Ziyun and Hua, Hang and Fu, Jianlong and Luo, Jiebo},
  booktitle={NeurIPS},
  year={2024}
}

@article{Painter,
  title={Images Speak in Images: A Generalist Painter for In-Context Visual Learning},
  author={Wang, Xinlong and Wang, Wen and Cao, Yue and Shen, Chunhua and Huang, Tiejun},
  journal={arXiv preprint arXiv:2212.02499},
  year={2022}
}

@article{jiang2023autodir,
  title={Autodir: Automatic all-in-one image restoration with latent diffusion},
  author={Jiang, Yitong and Zhang, Zhaoyang and Xue, Tianfan and Gu, Jinwei},
  journal={arXiv preprint arXiv:2310.10123},
  year={2023}
}

@article{luo2023controlling,
  title={Controlling Vision-Language Models for Universal Image Restoration},
  author={Luo, Ziwei and Gustafsson, Fredrik K and Zhao, Zheng and Sj{\"o}lund, Jens and Sch{\"o}n, Thomas B},
  journal={arXiv preprint arXiv:2310.01018},
  year={2023}
}

@inproceedings{Zamir2021MPRNet,
    title={Multi-Stage Progressive Image Restoration},
    author={Syed Waqas Zamir and Aditya Arora and Salman Khan and Munawar Hayat
            and Fahad Shahbaz Khan and Ming-Hsuan Yang and Ling Shao},
    booktitle={CVPR},
    year={2021}
}

@article{tu2022maxim,
  title={MAXIM: Multi-Axis MLP for Image Processing},
  author={Tu, Zhengzhong and Talebi, Hossein and Zhang, Han and Yang, Feng and Milanfar, Peyman and Bovik, Alan and Li, Yinxiao},
  journal={CVPR},
  year={2022},
}

@InProceedings{wang2021gfpgan,
    author = {Xintao Wang and Yu Li and Honglun Zhang and Ying Shan},
    title = {Towards Real-World Blind Face Restoration with Generative Facial Prior},
    booktitle={The IEEE Conference on Computer Vision and Pattern Recognition (CVPR)},
    year = {2021}
}

@inproceedings{zhou2022codeformer,
    author = {Zhou, Shangchen and Chan, Kelvin C.K. and Li, Chongyi and Loy, Chen Change},
    title = {Towards Robust Blind Face Restoration with Codebook Lookup TransFormer},
    booktitle = {NeurIPS},
    year = {2022}
}

@misc{flux2024,
    author={Black Forest Labs},
    title={FLUX},
    year={2024},
    howpublished={\url{https://github.com/black-forest-labs/flux}},
}

@misc{xie2024sana,
      title={Sana: Efficient High-Resolution Image Synthesis with Linear Diffusion Transformer},
      author={Enze Xie and Junsong Chen and Junyu Chen and Han Cai and Haotian Tang and Yujun Lin and Zhekai Zhang and Muyang Li and Ligeng Zhu and Yao Lu and Song Han},
      year={2024},
      eprint={2410.10629},
      archivePrefix={arXiv},
      primaryClass={cs.CV},
      url={https://arxiv.org/abs/2410.10629},
    }

@article{gao2024lumin-t2x,
  title={Lumina-T2X: Transforming Text into Any Modality, Resolution, and Duration via Flow-based Large Diffusion Transformers},
  author={Gao, Peng and Zhuo, Le and Liu, Chris and and Du, Ruoyi and Luo, Xu and Qiu, Longtian and Zhang, Yuhang and others},
  journal={arXiv preprint arXiv:2405.05945},
  year={2024}
}

@article{gao2024lumina-next,
  title={Lumina-Next: Making Lumina-T2X Stronger and Faster with Next-DiT},
  author={Zhuo, Le and Du, Ruoyi and Han, Xiao and Li, Yangguang and Liu, Dongyang and Huang, Rongjie and Liu, Wenze and others},
  journal={arXiv preprint arXiv:2406.18583},
  year={2024}
}

@inproceedings{huang2017arbitrary,
  title={Arbitrary style transfer in real-time with adaptive instance normalization},
  author={Huang, Xun and Belongie, Serge},
  booktitle={Proceedings of the IEEE international conference on computer vision},
  pages={1501--1510},
  year={2017}
}

@inproceedings{gatys2016image,
  title={Image style transfer using convolutional neural networks},
  author={Gatys, Leon A and Ecker, Alexander S and Bethge, Matthias},
  booktitle={Proceedings of the IEEE conference on computer vision and pattern recognition},
  pages={2414--2423},
  year={2016}
}

@inproceedings{esser2024scaling,
  title={Scaling rectified flow transformers for high-resolution image synthesis},
  author={Esser, Patrick and Kulal, Sumith and Blattmann, Andreas and Entezari, Rahim and M{\"u}ller, Jonas and Saini, Harry and Levi, Yam and Lorenz, Dominik and Sauer, Axel and Boesel, Frederic and others},
  booktitle={Forty-first international conference on machine learning},
  year={2024}
}

@inproceedings{rombach2022high,
  title={High-resolution image synthesis with latent diffusion models},
  author={Rombach, Robin and Blattmann, Andreas and Lorenz, Dominik and Esser, Patrick and Ommer, Bj{\"o}rn},
  booktitle={Proceedings of the IEEE/CVF conference on computer vision and pattern recognition},
  pages={10684--10695},
  year={2022}
}

@article{ho2020denoising,
  title={Denoising diffusion probabilistic models},
  author={Ho, Jonathan and Jain, Ajay and Abbeel, Pieter},
  journal={Advances in neural information processing systems},
  volume={33},
  pages={6840--6851},
  year={2020}
}

@inproceedings{guo2024onerestore,
  title={OneRestore: A Universal Restoration Framework for Composite Degradation},
  author={Guo, Yu and Gao, Yuan and Lu, Yuxu and Liu, Ryan Wen and He, Shengfeng},
  booktitle={European Conference on Computer Vision},
  year={2024}
}

@inproceedings{potlapalli2023promptir,
  title={PromptIR: Prompting for All-in-One Image Restoration},
  author={Potlapalli, Vaishnav and Zamir, Syed Waqas and Khan, Salman and Khan, Fahad},
  booktitle={Thirty-seventh Conference on Neural Information Processing Systems},
  year={2023}
}

@inproceedings{li2022all,
  title={All-in-one image restoration for unknown corruption},
  author={Li, Boyun and Liu, Xiao and Hu, Peng and Wu, Zhongqin and Lv, Jiancheng and Peng, Xi},
  booktitle={Proceedings of the IEEE/CVF conference on computer vision and pattern recognition},
  pages={17452--17462},
  year={2022}
}

@article{ravi2024sam,
  title={Sam 2: Segment anything in images and videos},
  author={Ravi, Nikhila and Gabeur, Valentin and Hu, Yuan-Ting and Hu, Ronghang and Ryali, Chaitanya and Ma, Tengyu and Khedr, Haitham and R{\"a}dle, Roman and Rolland, Chloe and Gustafson, Laura and others},
  journal={arXiv preprint arXiv:2408.00714},
  year={2024}
}

@inproceedings{kirillov2023segment,
  title={Segment anything},
  author={Kirillov, Alexander and Mintun, Eric and Ravi, Nikhila and Mao, Hanzi and Rolland, Chloe and Gustafson, Laura and Xiao, Tete and Whitehead, Spencer and Berg, Alexander C and Lo, Wan-Yen and others},
  booktitle={Proceedings of the IEEE/CVF international conference on computer vision},
  pages={4015--4026},
  year={2023}
}

@inproceedings{yang2024depth,
  title={Depth anything: Unleashing the power of large-scale unlabeled data},
  author={Yang, Lihe and Kang, Bingyi and Huang, Zilong and Xu, Xiaogang and Feng, Jiashi and Zhao, Hengshuang},
  booktitle={Proceedings of the IEEE/CVF Conference on Computer Vision and Pattern Recognition},
  pages={10371--10381},
  year={2024}
}

@inproceedings{chen2024comparative,
  title={A comparative study of image restoration networks for general backbone network design},
  author={Chen, Xiangyu and Li, Zheyuan and Pu, Yuandong and Liu, Yihao and Zhou, Jiantao and Qiao, Yu and Dong, Chao},
  booktitle={European Conference on Computer Vision},
  pages={74--91},
  year={2024},
  organization={Springer}
}

@article{zhang2017beyond,
  title={Beyond a gaussian denoiser: Residual learning of deep cnn for image denoising},
  author={Zhang, Kai and Zuo, Wangmeng and Chen, Yunjin and Meng, Deyu and Zhang, Lei},
  journal={IEEE transactions on image processing},
  volume={26},
  number={7},
  pages={3142--3155},
  year={2017},
  publisher={IEEE}
}

@article{dong2015image,
  title={Image super-resolution using deep convolutional networks},
  author={Dong, Chao and Loy, Chen Change and He, Kaiming and Tang, Xiaoou},
  journal={IEEE transactions on pattern analysis and machine intelligence},
  volume={38},
  number={2},
  pages={295--307},
  year={2015},
  publisher={IEEE}
}

@inproceedings{chen2023activating,
  title={Activating more pixels in image super-resolution transformer},
  author={Chen, Xiangyu and Wang, Xintao and Zhou, Jiantao and Qiao, Yu and Dong, Chao},
  booktitle={Proceedings of the IEEE/CVF conference on computer vision and pattern recognition},
  pages={22367--22377},
  year={2023}
}

@article{chen2024expanding,
  title={Expanding Performance Boundaries of Open-Source Multimodal Models with Model, Data, and Test-Time Scaling},
  author={Chen, Zhe and Wang, Weiyun and Cao, Yue and Liu, Yangzhou and Gao, Zhangwei and Cui, Erfei and Zhu, Jinguo and Ye, Shenglong and Tian, Hao and Liu, Zhaoyang and others},
  journal={arXiv preprint arXiv:2412.05271},
  year={2024}
}

@article{alayrac2022flamingo,
  title={Flamingo: a visual language model for few-shot learning},
  author={Alayrac, Jean-Baptiste and Donahue, Jeff and Luc, Pauline and Miech, Antoine and Barr, Iain and Hasson, Yana and Lenc, Karel and Mensch, Arthur and Millican, Katherine and Reynolds, Malcolm and others},
  journal={Advances in neural information processing systems},
  volume={35},
  pages={23716--23736},
  year={2022}
}

@article{chen2024far,
  title={How far are we to gpt-4v? closing the gap to commercial multimodal models with open-source suites},
  author={Chen, Zhe and Wang, Weiyun and Tian, Hao and Ye, Shenglong and Gao, Zhangwei and Cui, Erfei and Tong, Wenwen and Hu, Kongzhi and Luo, Jiapeng and Ma, Zheng and others},
  journal={Science China Information Sciences},
  volume={67},
  number={12},
  pages={220101},
  year={2024},
  publisher={Springer}
}

@inproceedings{chen2024internvl,
  title={Internvl: Scaling up vision foundation models and aligning for generic visual-linguistic tasks},
  author={Chen, Zhe and Wu, Jiannan and Wang, Wenhai and Su, Weijie and Chen, Guo and Xing, Sen and Zhong, Muyan and Zhang, Qinglong and Zhu, Xizhou and Lu, Lewei and others},
  booktitle={Proceedings of the IEEE/CVF Conference on Computer Vision and Pattern Recognition},
  pages={24185--24198},
  year={2024}
}

@article{achiam2023gpt,
  title={Gpt-4 technical report},
  author={Achiam, Josh and Adler, Steven and Agarwal, Sandhini and Ahmad, Lama and Akkaya, Ilge and Aleman, Florencia Leoni and Almeida, Diogo and Altenschmidt, Janko and Altman, Sam and Anadkat, Shyamal and others},
  journal={arXiv preprint arXiv:2303.08774},
  year={2023}
}

@article{chen2024unireal,
  title={UniReal: Universal Image Generation and Editing via Learning Real-world Dynamics},
  author={Chen, Xi and Zhang, Zhifei and Zhang, He and Zhou, Yuqian and Kim, Soo Ye and Liu, Qing and Li, Yijun and Zhang, Jianming and Zhao, Nanxuan and Wang, Yilin and others},
  journal={arXiv preprint arXiv:2412.07774},
  year={2024}
}

@article{liu2023unifying,
  title={Unifying image processing as visual prompting question answering},
  author={Liu, Yihao and Chen, Xiangyu and Ma, Xianzheng and Wang, Xintao and Zhou, Jiantao and Qiao, Yu and Dong, Chao},
  journal={arXiv preprint arXiv:2310.10513},
  year={2023}
}

@inproceedings{chen2024learning,
  title={Learning a low-level vision generalist via visual task prompt},
  author={Chen, Xiangyu and Liu, Yihao and Pu, Yuandong and Zhang, Wenlong and Zhou, Jiantao and Qiao, Yu and Dong, Chao},
  booktitle={Proceedings of the 32nd ACM International Conference on Multimedia},
  pages={2671--2680},
  year={2024}
}

@article{han2024ace,
  title={ACE: All-round Creator and Editor Following Instructions via Diffusion Transformer},
  author={Han, Zhen and Jiang, Zeyinzi and Pan, Yulin and Zhang, Jingfeng and Mao, Chaojie and Xie, Chenwei and Liu, Yu and Zhou, Jingren},
  journal={arXiv preprint arXiv:2410.00086},
  year={2024}
}

@article{xiao2024omnigen,
  title={Omnigen: Unified image generation},
  author={Xiao, Shitao and Wang, Yueze and Zhou, Junjie and Yuan, Huaying and Xing, Xingrun and Yan, Ruiran and Wang, Shuting and Huang, Tiejun and Liu, Zheng},
  journal={arXiv preprint arXiv:2409.11340},
  year={2024}
}

@article{wang2024lavin,
  title={Lavin-dit: Large vision diffusion transformer},
  author={Wang, Zhaoqing and Xia, Xiaobo and Chen, Runnan and Yu, Dongdong and Wang, Changhu and Gong, Mingming and Liu, Tongliang},
  journal={arXiv preprint arXiv:2411.11505},
  year={2024}
}

@article{wang2023context,
  title={In-context learning unlocked for diffusion models},
  author={Wang, Zhendong and Jiang, Yifan and Lu, Yadong and He, Pengcheng and Chen, Weizhu and Wang, Zhangyang and Zhou, Mingyuan and others},
  journal={Advances in Neural Information Processing Systems},
  volume={36},
  pages={8542--8562},
  year={2023}
}

@inproceedings{mou2024t2i,
  title={T2i-adapter: Learning adapters to dig out more controllable ability for text-to-image diffusion models},
  author={Mou, Chong and Wang, Xintao and Xie, Liangbin and Wu, Yanze and Zhang, Jian and Qi, Zhongang and Shan, Ying},
  booktitle={Proceedings of the AAAI Conference on Artificial Intelligence},
  volume={38},
  number={5},
  pages={4296--4304},
  year={2024}
}

@inproceedings{zhang2023adding,
  title={Adding conditional control to text-to-image diffusion models},
  author={Zhang, Lvmin and Rao, Anyi and Agrawala, Maneesh},
  booktitle={Proceedings of the IEEE/CVF International Conference on Computer Vision},
  pages={3836--3847},
  year={2023}
}

@article{wang2024emu3,
  title={Emu3: Next-token prediction is all you need},
  author={Wang, Xinlong and Zhang, Xiaosong and Luo, Zhengxiong and Sun, Quan and Cui, Yufeng and Wang, Jinsheng and Zhang, Fan and Wang, Yueze and Li, Zhen and Yu, Qiying and others},
  journal={arXiv preprint arXiv:2409.18869},
  year={2024}
}

@misc{le2024diffusiongenerate,
      title={One Diffusion to Generate Them All}, 
      author={Duong H. Le and Tuan Pham and Sangho Lee and Christopher Clark and Aniruddha Kembhavi and Stephan Mandt and Ranjay Krishna and Jiasen Lu},
      year={2024},
      eprint={2411.16318},
      archivePrefix={arXiv},
      primaryClass={cs.CV},
      url={https://arxiv.org/abs/2411.16318}, 
}

@article{mao2025ace++,
  title={ACE++: Instruction-Based Image Creation and Editing via Context-Aware Content Filling},
  author={Mao, Chaojie and Zhang, Jingfeng and Pan, Yulin and Jiang, Zeyinzi and Han, Zhen and Liu, Yu and Zhou, Jingren},
  journal={arXiv preprint arXiv:2501.02487},
  year={2025}
}

@InProceedings{lolV2,
author = {Yang, Wenhan and Wang, Shiqi and Fang, Yuming and Wang, Yue and Liu, Jiaying},
title = {From Fidelity to Perceptual Quality: A Semi-Supervised Approach for Low-Light Image Enhancement},
booktitle = {IEEE/CVF Conference on Computer Vision and Pattern Recognition (CVPR)},
month = {June},
year = {2020}
}

@inproceedings{mit5k,
	author = "Vladimir Bychkovsky and Sylvain Paris and Eric Chan and Fr{\'e}do Durand",
	title = "Learning Photographic Global Tonal Adjustment with a Database of Input / Output Image Pairs",
	booktitle = "The Twenty-Fourth IEEE Conference on Computer Vision and Pattern Recognition",
	year = "2011"
}

@article{chen2025exploring,
  title={Exploring Scalable Unified Modeling for General Low-Level Vision},
  author={Chen, Xiangyu and Zhu, Kaiwen and Pu, Yuandong and Cao, Shuo and Li, Xiaohui and Zhang, Wenlong and Liu, Yihao and Qiao, Yu and Zhou, Jiantao and Dong, Chao},
  journal={arXiv preprint arXiv:2507.14801},
  year={2025}
}

@article{hypir,
  title={Harnessing Diffusion-Yielded Score Priors for Image Restoration},
  author={Lin, Xinqi and Yu, Fanghua and Hu, Jinfan and You, Zhiyuan and Shi, Wu and Ren, Jimmy S and Gu, Jinjin and Dong, Chao},
  journal={arXiv preprint arXiv:2507.20590},
  year={2025}
}
